\documentclass[conference]{IEEEtran}
\IEEEoverridecommandlockouts

\usepackage[utf8]{inputenc}
\usepackage{amsmath}
\usepackage{amsfonts}
\usepackage{amssymb}
\usepackage{amsthm}
\usepackage{mathtools}
\usepackage{tikz}

\usepackage{graphicx}
\usepackage{subcaption}
\usepackage[export]{adjustbox}
\usepackage{float}  

\usepackage{array}
\usepackage{booktabs}
\usepackage{multirow}

\usepackage{algorithm}
\usepackage{algpseudocode}

\usepackage{hyperref}
\usepackage{xcolor}

\title{Polar Separable Transform for Efficient Orthogonal Rotation-Invariant Image Representation}

\author{Satya P. Singh,~\IEEEmembership{Member,~IEEE,}
        Rashmi Chaudhry,
        Anand Srivastava,~and~Jagath C. Rajapakse,~\IEEEmembership{Fellow,~IEEE}
\thanks{S. P. Singh is with the Department of Electronics and Communication Engineering, Netaji Subhas University of Technology (NSUT), New Delhi, India, e-mail: satyapsingh@nsut.ac.in.}%
\thanks{R. Chaudhry is with the Department of Computer Science and Engineering, Netaji Subhas University of Technology (NSUT), New Delhi, India, e-mail: rashmic@nsut.ac.in.}%
\thanks{A. Srivastava is the Vice Chancellor of Netaji Subhas University of Technology (NSUT), New Delhi, India, e-mail: vc@nsut.ac.in.}%
\thanks{J. C. Rajapakse is with the College of Computing and Data Science, Nanyang Technological University (NTU), Singapore, e-mail: ASJagath@ntu.edu.sg.}%
}

\begin{document}

\maketitle
\begin{tikzpicture}[remember picture,overlay]
\node at ([yshift=1em]current page.south) {\parbox{\textwidth}{\centering
\small \textbf{This work has been submitted to the IEEE for possible publication. 
Copyright may be transferred without notice, after which this version may no longer be accessible.}}};
\end{tikzpicture}

\begin{abstract}
Orthogonal moment-based image representations are fundamental in computer vision, but classical methods suffer from high computational complexity and numerical instability at large orders. Zernike and pseudo-Zernike moments, for instance, require coupled radial-angular processing that precludes efficient factorization, resulting in $\mathcal{O}(n^3N^2)$ to $\mathcal{O}(n^6N^2)$ complexity and $\mathcal{O}(N^4)$ condition number scaling for the $n$th-order moments on an $N\times N$ image. We introduce \textbf{PSepT} (Polar Separable Transform), a separable orthogonal transform that overcomes the non-separability barrier in polar coordinates. PSepT achieves complete kernel factorization via tensor-product construction of Discrete Cosine Transform (DCT) radial bases and Fourier harmonic angular bases, enabling independent radial and angular processing. This separable design reduces computational complexity to $\mathcal{O}(N^2 \log N)$, memory requirements to $\mathcal{O}(N^2)$, and condition number scaling to $\mathcal{O}(\sqrt{N})$, representing exponential improvements over polynomial approaches. PSepT exhibits orthogonality, completeness, energy conservation, and rotation-covariance properties. Experimental results demonstrate better numerical stability, computational efficiency, and competitive classification performance on structured datasets, while preserving exact reconstruction. The separable framework enables high-order moment analysis previously infeasible with classical methods, opening new possibilities for robust image analysis applications.
\end{abstract}

\section{Introduction}
\IEEEPARstart{O}{rthogonal} moment-based image representation has emerged as a fundamental technique in computer vision and pattern recognition due to its attracticve mathematical properties, particularly geometric invariance and independence \cite{wang2021survey}. The field has evolved through several distinct phases, each addressing specific limitations while facing new computational challenges. Teague laid the foundation with the introduction of orthogonal moments based on continuous polynomial functions \cite{teague1980image}. Since then, moment-based methods have found wide applications in character recognition, face recognition, direction-of-arrival estimation, and trademark segmentation \cite{singh2014survey}. Among these, Zernike moments (ZM) \cite{Khotanzad1990} and pseudo-Zernike moments (PZM) \cite{chong2003scale}, defined using complex polynomials over the unit disk, became widely adopted due to their natural rotation invariance \cite{khotanzad1990invariant,liao1998accuracy}. Nevertheless, the computational burden associated with high-order polynomial evaluation has remained a persistent challenge.

Recognition of numerical instabilities in continuous-to-discrete mapping led to the development of discrete orthogonal moments. Methods such as discrete Tchebichef, Krawtchouk, and Hahn moments were introduced to address these issues \cite{mukundan2004computational,mukundan1998moment}. Racah polynomials were introduced into image analysis to demonstrate their potential usefulness, with scaling applied for maintaining numerical stability within the range [-1, 1] \cite{zhu2007image}. While these approaches improved numerical conditioning, they suffered from fundamental computational complexity limitations. Recent developments introduced polar harmonic transforms (PHT) including Polar Cosine Transform (PCT), Polar Sine Transform (PST), and Polar Complex Exponential Transform (PCET) \cite{yap2009two}. These methods attempted to balance computational efficiency with mathematical rigor, achieving moderate improvements over classical polynomial approaches while maintaining some degree of orthogonality.

Despite extensive research into computation schemes for efficient calculation of orthogonal image moments, comparative studies reveal that each scheme faces fundamental limitations under different conditions \cite{papakostas2010computation}. The computational complexity of traditional orthogonal moments scales prohibitively with the moment order and image size. Classical methods require direct evaluation of high-order polynomials at each image pixel, resulting in $O(n^3N^2)$ to $O(n^6N^2)$ complexity for $n$-th order moments on $N \times N$ images. This complexity barrier has limited practical applications to low-order moments, restricting the representational capacity of these methods. Traditional approaches require storing full kernel matrices, leading to $O(n^2N^2)$ memory complexity that becomes prohibitive for high-resolution images or high-order analysis \cite{xin2010circularly,ren2003multidistortion,yap2003image}. The coupled nature of polynomial kernels prevents efficient factorization or compression of these representations. Usage of continuous moments for discrete images brings about certain problems in terms of numerical stability \cite{mukundan2009orthogonal}. High-order polynomial moments exhibit condition numbers scaling as $O(N^4)$, leading to catastrophic numerical failures that render these methods impractical for detailed analysis. Limited work has explored separability in related contexts. Research on separable versus non-separable transforms in video coding has shown that while non-separable transforms can achieve better compression performance, separable transforms offer significant computational advantages \cite{chen2018joint}. However, achieving separability in polar coordinates while maintaining orthogonality has remained an unsolved problem.

Recent developments in two-dimensional Discrete Fourier Transform (DFT) in polar coordinates have demonstrated computational advantages, but no discrete counterpart exists for general orthogonal moments \cite{zhao2021discrete1,zhao2021discrete2}. Obtaining exact and fast discrete Fourier transforms on polar grids remains challenging, with no known exact solutions \cite{averbuch2020exact}. The discrete cosine transform (DCT), introduced by Nasir Ahmed in 1972, is widely used in signal processing and data compression \cite{ahmed1974discrete}. DCT-II is asymptotically equivalent to the Karhunen-Loève transform for Markov signals, making it near-optimal for smooth signals \cite{rao2014discrete}, while Fast Fourier Transform (FFT) algorithms achieve $O(N \log N)$ complexity by factorizing the DFT matrix \cite{cooley1965algorithm}.  

Linear, orthogonal, separable transforms are popular for their energy-preserving properties \cite{rao2014discrete}. Separable architectures enable independent processing of different dimensions, reducing computational complexity via factorization. Despite extensive theoretical development, rotation invariance of orthogonal moments is limited in practice due to computational constraints \cite{hoang2021survey}, preventing scalable high-order analysis required for complex pattern recognition.  

Existing polar transforms, such as Zernike and pseudo-Zernike moments, provide orthogonal bases on the unit disk but exhibit inherent radial-angular coupling. This complicates the design of compact, fully separable descriptors, creating a trade-off between robustness and discriminative power, particularly with a limited set of stable features. High-order moment extraction is further constrained by factorial growth in radial polynomials and recursive dependencies, leading to algorithms whose complexity scales steeply with moment order \cite{zhang2024perceptual}.  

To address these limitations, we introduce PSepT (Polar Separable Transform) that employs a tensor product of DCT radial bases and Fourier harmonic angular bases, enabling independent radial-angular processing. This reduces computational complexity from $\mathcal{O}(N^3)$ to $\mathcal{O}(N^2 \log N)$, lowers memory requirements, and improves numerical stability, representing a significant advancement over conventional methods. The main contributions of this paper are summarized as follows:

\begin{itemize}
    \item We introduce PSepT, a separable tensor-product polar transform with DCT-II radial and Fourier angular bases, which achieves strict discrete orthogonality and efficient two-stage forward/inverse algorithms of complexity $\mathcal{O}(N_r N_\theta \log N)$.
    \item The framework provides rigorous numerical analysis, a stable polar sampling grid, and establishes rotation covariance where coefficient magnitudes yield natural invariants.
    \item Extensive experiments on standard datasets demonstrate PSepT’s highnumerical stability, fast run times, and strong rotation robustness, while noting trade-offs in representing highly textured natural images.
\end{itemize}

The rest of this paper is organized as follows. Section~\ref{2} presents the mathematical foundation and tensor product decomposition while Section~\ref{3} introduces the PSepT kernel and proves its discrete orthogonality. Sections~\ref{4} and~\ref{5} describe forward and inverse transforms, highlighting the two-stage DCT-FFT framework for efficient computation. Section~\ref{6} analyzes properties including energy conservation, rotation covariance, and convergence. Section~\ref{7} reports experiments on reconstruction, numerical stability, rotation invariance, and classification. Section~\ref{8} concludes with a summary of contributions, practical benefits, computational advantages, limitations, and future directions.

\section{Mathematical Foundations: Separable Kernel Theory} \label{2}

Let $\mathcal{F} = L^2(\Omega)$ be the space of square-integrable functions defined on the image domain $\Omega \subset \mathbb{R}^2$, equipped with the inner product \cite{medoff1985inner}:
\begin{equation}
\langle f, g \rangle = \int_\Omega f(x,y) g^*(x,y) \, dx \, dy,
\end{equation}
with induced norm $\|f\|_2 = \sqrt{\langle f, f \rangle}$.

A kernel function $K(x,y)$ is \textit{separable} if it can be expressed as
\begin{equation}
K(x,y) = \sum_{i} \alpha_i R_i(x) A_i(y),
\end{equation}
for some finite sets of functions, $\{R_i\}$ and $\{A_i\}$.

Define the polar coordinate transformation $T: \Omega \to D$ by
\begin{align}
T(x,y) &= (r, \theta), \\
r &= \sqrt{x^2 + y^2}, \\
\theta &= \arctan_2(y,x),
\end{align}
where $D = \{(r,\theta) : 0 \leq r \leq 1, -\pi \leq \theta < \pi\}$ is the unit disk in polar coordinates.  
The Jacobian of this transformation is $J(r,\theta) = r$, giving the measure \cite{folland1999real}:
\begin{equation}
\int_\Omega f(x,y) \, dx \, dy 
= \int_D f(r\cos\theta, r\sin\theta)\, r \, dr \, d\theta.
\end{equation}
A polar kernel $K(r,\theta)$ is \textit{polar-separable} if it can be written as
\begin{equation}
K(r,\theta) = R(r) \, A(\theta).
\end{equation}

Classical orthogonal polynomials defined on the unit disk, such as Zernike and pseudo-Zernike moments \cite{chong2003scale} cannot be expressed in this separable form while preserving orthogonality. For example, Zernike polynomials are defined as
\begin{equation}
Z_n^m(r,\theta) = R_n^{|m|}(r) e^{im\theta},
\end{equation}
where the radial polynomial $R_n^{|m|}(r)$ depends jointly on both indices $n$ and $m$. This coupling prevents factorization into independent radial and angular components while maintaining orthogonality \cite{Liao1998}:
\begin{equation}
\int_0^{2\pi} \int_0^1 
Z_n^m(r,\theta) Z_{n'}^{m'}(r,\theta)^* \, r \, dr \, d\theta 
= \delta_{n,n'} \delta_{m,m'}.
\end{equation}
Thus, the interdependence of $n$ and $m$ in the radial functions makes a strictly separable decomposition impossible.

\section{PSepT Separable Kernel Architecture} \label{3}

Separable orthogonal kernels in polar coordinates are formulated using discrete transform theory, overcoming coupled radial-angular limitations of polynomial methods and enabling efficient, orthogonal radial basis computation \cite{ahmed1974discrete}:
\begin{equation}
\small
\Phi_n(r_k) = \sqrt{\frac{2}{N_r}} \cos\Big(\frac{n \pi (k + 1/2)}{N_r}\Big), \quad n = 0,1,\ldots,N_r-1
\end{equation}

The functions $\Phi_n(r_k)$ form an orthogonal basis over the discrete radial grid, ensuring rigorous and comprehensive representation. Their structure enables efficient computation with complexity $O(N \log N)$ via optimized fast Discrete Cosine Transform (DCT) routines. The radial basis achieves near-optimal energy compaction for smoothly varying radial profiles while maintaining high numerical stability, with a condition number scaling asymptotically as $O(\sqrt{N})$. The discrete orthogonality, supported by \cite{rao2014discrete,cooley1965algorithm}, is expressed as:
\begin{equation}
\sum_{k=0}^{N_r-1} \Phi_n(r_k) \Phi_{n'}(r_k) = \delta_{n,n'}
\end{equation}

 In the angular domain, we use angular basis functions based on Fourier harmonics for optimal representation. \cite{zhao2021discrete2}:
\begin{equation}
\Psi_m(\theta_j) = \frac{1}{\sqrt{N_\theta}} e^{im\theta_j}, \quad m \in \{-N_\theta/2, \ldots, N_\theta/2-1\}
\end{equation}
The basis functions is implemented using the FFT with computational complexity of $O(N \log N)$. These functions are designed to ensure natural rotation covariance and maintain perfect numerical conditioning with condition number $\kappa = 1$. The orthogonality property of the angular components is given by
\begin{equation}
\sum_{j=0}^{N_\theta-1} \Psi_m(\theta_j)\Psi_{m'}^*(\theta_j) = \delta_{m,m'}
\end{equation}

The separable kernel is constructed using a tensor product architecture, resulting in an orthogonal kernel in polar coordinates, which factors completely into independent radial and angular components:
\begin{align}
K_{n,m}(r_k, \theta_j)
  &= \Phi_n(r_k) \times \Psi_m(\theta_j) \notag \\
  &= \sqrt{\frac{2}{N_r N_\theta}} 
     \cos\!\left(\frac{n\pi(k + 1/2)}{N_r}\right) 
     e^{im\theta_j}
\end{align}
The separable kernel architecture preserves complete orthogonality through the fundamental tensor-product relationship 
 \cite{hackbusch2012tensor}:
\begin{equation}
\sum_{k=0}^{N_r-1} \sum_{j=0}^{N_\theta-1} K_{n,m}(r_k, \theta_j) K_{n',m'}^*(r_k, \theta_j) = \delta_{n,n'} \delta_{m,m'}
\end{equation}

\begin{proof}
The structure enables factorization of the orthogonality computation:
\begin{align*}
&\sum_{k=0}^{N_r-1} \sum_{j=0}^{N_\theta-1} K_{n,m}(r_k, \theta_j) K_{n',m'}^*(r_k, \theta_j) \\
&= \sum_{k=0}^{N_r-1} \sum_{j=0}^{N_\theta-1} \Phi_n(r_k) \Psi_m(\theta_j)\Phi_{n'}(r_k) \Psi_{m'}^*(\theta_j) \\
&= \left[\sum_{k=0}^{N_r-1} Phi_n(r_k) \Phi_{n'}(r_k)\right] \left[\sum_{j=0}^{N_\theta-1} \Psi_m(\theta_j) \Psi_{m'}^*(\theta_j)\right] \\
&= \delta_{n,n'} \delta_{m,m'}
\end{align*}

\end{proof}

To achieve separability, a specially designed discrete polar grid that preserves the factorizable structure required for the kernel architecture is constructed \cite{averbuch2008framework,potts1998fast}:
\begin{align}
G_{N_r,N_\theta}
   &= \Bigl\{ (r_k, \theta_j) : 
      r_k = \tfrac{k}{N_r-1} \cdot R_{\max}, \notag \\
   &\quad \quad \quad \quad\quad\quad \theta_j = -\pi + \tfrac{2\pi j}{N_\theta} \Bigr\}
\end{align}
where $k \in \{0,1,\ldots,N_r-1\}$, $j \in \{0,1,\ldots,N_\theta-1\}$, and $R_{\max} = 0.999$ is chosen to avoid singularities at the boundary. 

The construction of the discrete grid preserves the separability properties necessary for the factorization of the kernel \cite{potts1998fast}. Specifically, if a continuous kernel exhibits separability in the form $K(r,\theta) = R(r)A(\theta)$, then the evaluation of the discrete grid maintains this structure as $K[k,j] = R[k] \times A[j]$. 

\begin{figure}[h!]
    \centering
    \begin{subfigure}{0.9\linewidth}
        \centering
        \includegraphics[width=\linewidth]{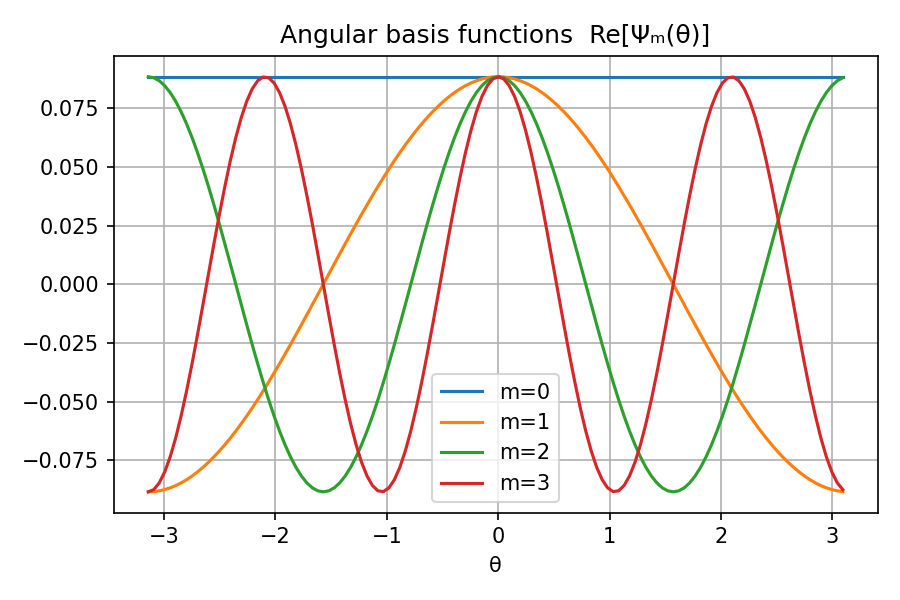}
        \caption{Angular basis functions $\Psi_{m}(\theta)$}
        \label{fig:pst_angular_basis}
    \end{subfigure}
    
    \vspace{1em} 
    
    \begin{subfigure}{0.9\linewidth}
        \centering
        \includegraphics[width=\linewidth]{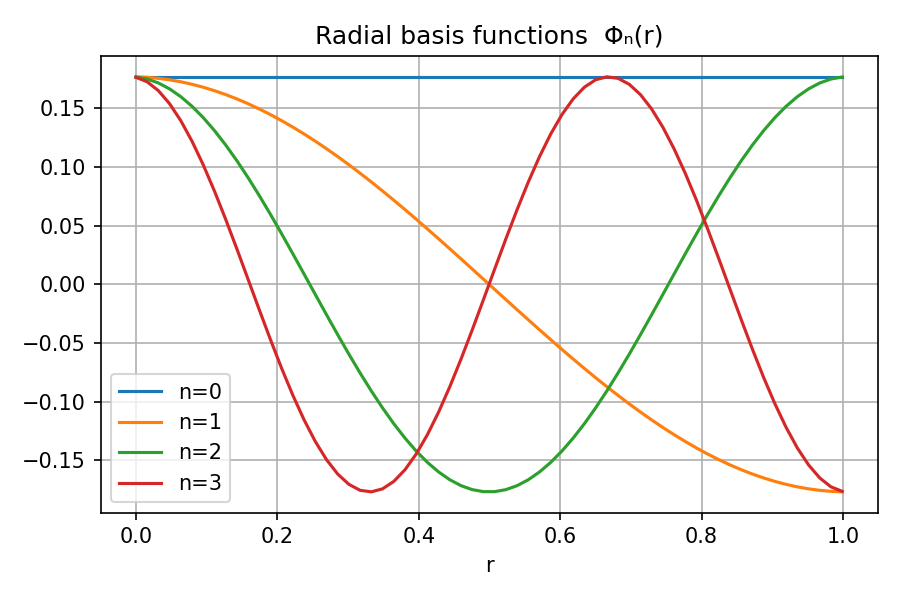}
        \caption{Radial basis functions $\Phi_{n}(r)$}
        \label{fig:pst_radial_basis}
    \end{subfigure}

    \caption{Separable components of PSepT kernels. (a) shows radial basis functions $\Phi_{n}(r)$ and (b) shows angular basis functions $\Psi_{m}(\theta)$ for orders $n,m=0,1,2,3$.}
    \label{fig:pst_components}
\end{figure}

Figure~\ref{fig:pst_components} illustrates the separable components that form the foundation of PSepT kernels, showing the radial basis functions $\Phi_{n}(r)$ (a) and angular basis functions $\Psi_{m}(\theta)$ (b) for orders $n,m=0,1,2,3$. The radial components exhibit the characteristic DCT-II cosine patterns with increasing oscillation frequency from center to edge, while the angular components display harmonic variations around the unit circle. Critically, these components combine through multiplication in the tensor product framework \cite{reed1980methods}:
\begin{equation}
    K_{n,m}(r,\theta) = \Phi_{n}(r) \times \Psi_{m}(\theta),
    \label{tensor}
\end{equation}
which enables separable processing and distinguishes the PSepT from conventional polynomial-based approaches.

\begin{figure}[h!]
    \centering
    \includegraphics[width=0.95\linewidth]{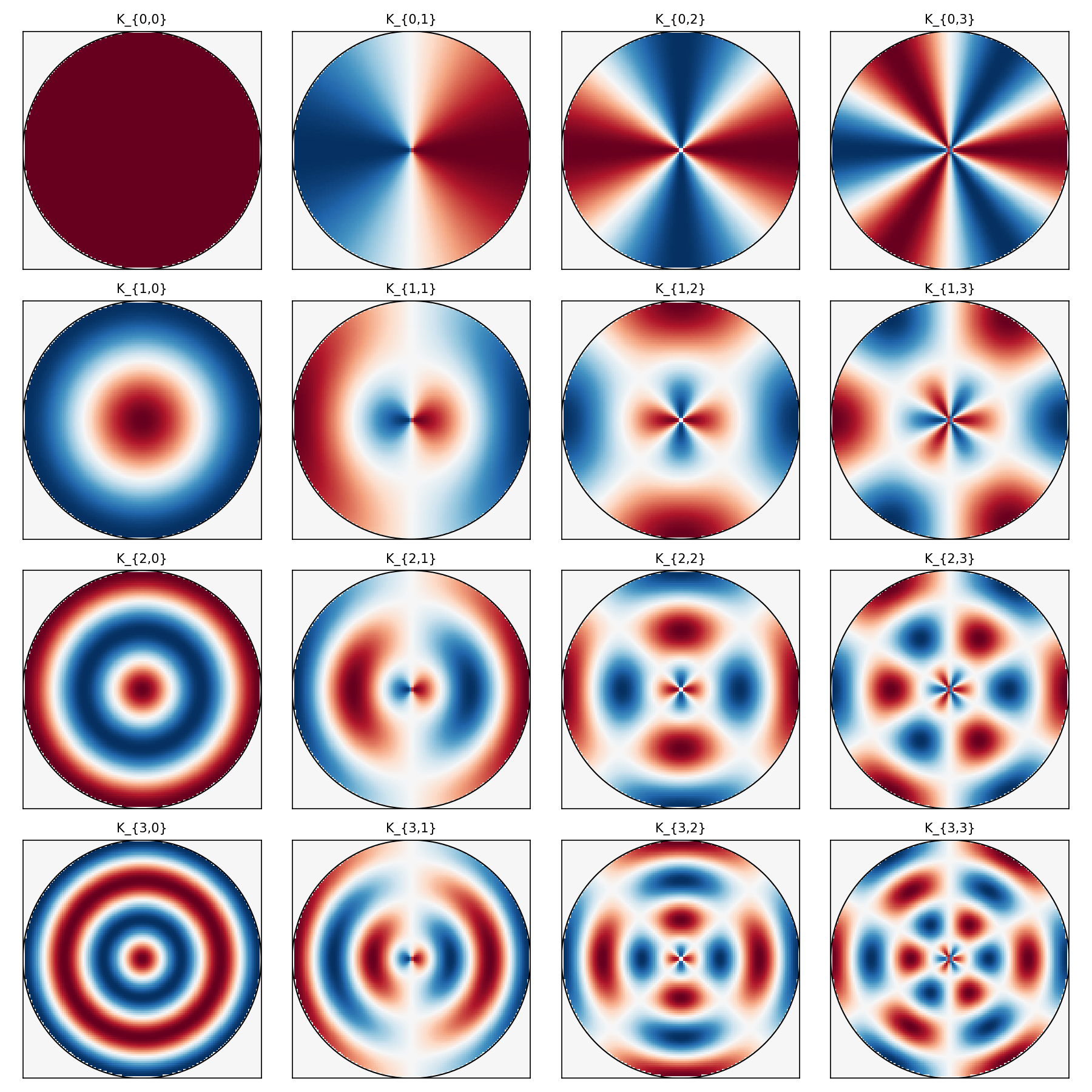}
    \caption{Comprehensive kernel gallery showing PSepT kernels $K_{n,m}$ for $n,m=0\text{--}4$ . Rows correspond to radial complexity ($n$), and columns correspond to angular complexity ($m$).}
    \label{fig:pst_gallery}
\end{figure}

\begin{figure}[h!]
    \centering
    \includegraphics[width=0.85\linewidth]{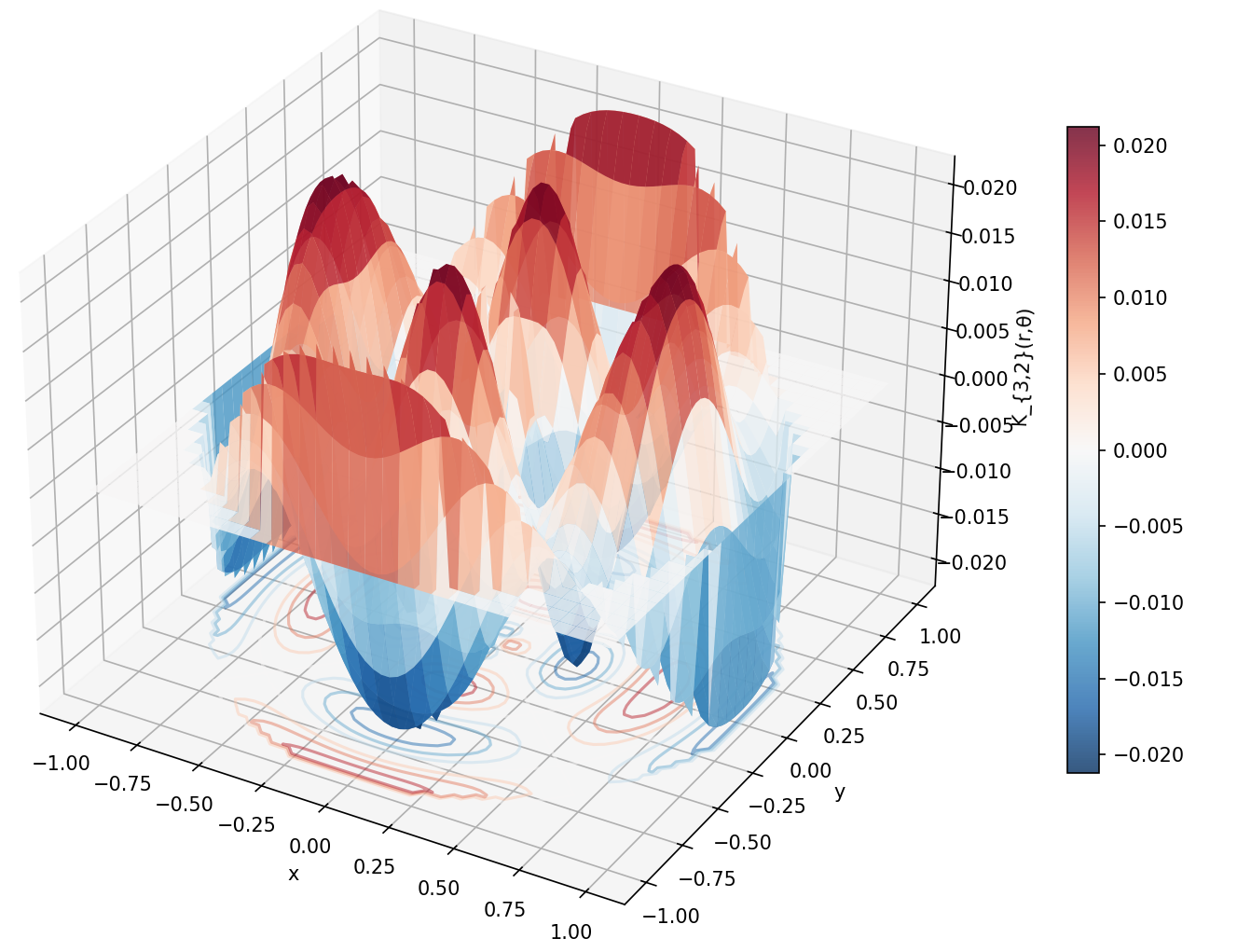}
    \caption{Three-dimensional visualization of kernel $K_{3,2}$ over the unit disk, showing three radial oscillations modulated by two angular cycles.}
    \label{fig:pst_kernel_3d}
\end{figure}

Figure~\ref{fig:pst_gallery} presents a comprehensive kernel gallery showing complete PSepT kernels $K_{n,m}$ for $n,m=0\text{--}4$ , clearly demonstrating how radial complexity (rows) and angular complexity (columns) combine independently. The center column ($m=0$) shows purely radial patterns with rotational symmetry, while increasing $|m|$ values introduce angular sectoring.  Figure~\ref{fig:pst_kernel_3d} provides a three-dimensional surface visualization of kernel $K_{3,2}$, showing the mathematical structure over the unit disk with three radial oscillations modulated by two angular cycles, resulting in the characteristic pattern with four main lobes.

\section{PSepT Forward Transform} \label{4}

\subsection{Separability-Enabled Algorithm}

Our separable kernel structure enables a two-stage algorithm that would not be possible with traditional non-separable approaches. Given a discrete polar image $g[j,k] = g(\theta_j, r_k)$, we compute the PSepT coefficients through \cite{qi2021survey}:
\begin{equation}
C_{n,m} = \frac{1}{N_r N_\theta} \sum_{k=0}^{N_r-1} \sum_{j=0}^{N_\theta-1} g[j,k] K_{n,m}^*(r_k, \theta_j) w_k
\end{equation}
where $w_k = r_k$ accounts for the polar Jacobian. The separability property allows this computation to be decomposed into two independent stages, thereby modifying the overall algorithmic complexity.

\subsubsection{Stage 1: Separable Radial Processing}
Since the kernel exhibits a separable structure, $K_{n,m}(r_k, \theta_j) = \Phi_n(r_k) \times \Psi_m(\theta_j)$, each angular slice is processed independently. The radial DCT analysis is expressed as \cite{qi2021survey}:
\begin{equation}
G_{\text{DCT}}[j,n] = \sum_{k=0}^{N_r-1} g[j,k]\Phi_n(r_k)
\end{equation}
The radial processing stage exploits separability to achieve several advantages. Each angular slice $j$ is processed independently, enabling parallel computation and avoiding coupling between angular positions. The use of a DCT basis provides near-optimal energy compaction for smooth radial profiles, concentrating most of the information in low-order coefficients. Computation is efficient, with complexity $O(N_r \log N_r)$ per slice due to optimized DCT algorithms. In addition, the transformations exhibit favorable numerical stability with condition number $\kappa = O(\sqrt{N_r})$, which offers an improvement in conditioning compared to polynomial-based methods.

\subsubsection{Stage 2: Separable Angular Processing}
The separable structure allows independent angular processing of each radial mode, thereby avoiding coupling effects that occur in traditional approaches. The angular FFT analysis is given by \cite{qi2021survey}:
\begin{equation}
C_{n,m} = \sum_{j=0}^{N_\theta-1} G_{\text{DCT}}[j,n] e^{-i2\pi mj/N_\theta}
\end{equation}

The angular stage leverages separability by processing each radial mode $n$ independently, enabling parallel computation. FFT-based formulation captures all angular frequencies up to the Nyquist limit with complexity $O(N_\theta \log N_\theta)$ per mode. Perfect conditioning ($\kappa = 1$) avoids numerical instabilities typical of polynomial-based methods.

\begin{algorithm}
\caption{PSepT Separable Forward Transform}
\begin{algorithmic}[1]
\Require{Image $f(x,y)$, transform orders $(n_{\max}, m_{\max})$}
\Ensure{PSepT coefficients $\{C_{n,m}\}$}
\State
\State \textbf{Preprocessing:}
\State Transform $f(x,y)$ to polar coordinates $\rightarrow g(r,\theta)$
\State Generate separable polar grid $(r_k, \theta_j)$ with dimensions $(N_r, N_\theta)$
\State
\State \textbf{Stage 1: Radial Processing}
\For{each angular slice $j = 0, \ldots, N_\theta-1$}
    \State $G_{\text{DCT}}[j,:] \leftarrow \text{DCT}(g[j,:])$ \Comment{Apply DCT to radial profile}
\EndFor
\State
\State \textbf{Stage 2: Angular Processing}
\For{each radial mode $n = 0, \ldots, N_r-1$}
    \State $\{C_{n,m}\} \leftarrow \text{FFT}(G_{\text{DCT}}[:,n])$ \Comment{Apply FFT to angular data}
\EndFor
\State
\State \textbf{Coefficient Extraction:}
\State Extract $C_{n,m}$ for $n \leq n_{\max}$ and $|m| \leq m_{\max}$
\State
\Return{Separable coefficients $\{C_{n,m}\}$}
\end{algorithmic}
\end{algorithm}

\section{PSepT Inverse Transform} \label{5}

\subsection{Separable Reconstruction Theory}
The separable architecture of the PSepT enables reconstruction through a two-stage inverse process that mirrors the forward transform. For any finite coefficient set \(\{C_{n,m} : 0 \leq n < N_r, |m| < N_\theta/2\}\), the PSepT separable inverse achieves exact reconstruction. The reconstruction formula is given by \cite{katznelson2004introduction}:
\begin{equation}
g[j,k] = \sum_{n=0}^{N_r-1} \sum_{m=-N_\theta/2}^{N_\theta/2-1} C_{n,m} K_{n,m}(r_k, \theta_j)
\end{equation}
where the separable structure \(K_{n,m}(r_k, \theta_j) = \Phi_n(r_k) \times \Psi_m(\theta_j)\) allows reconstruction to be computed in two independent stages. This property facilitates efficient computation of the inverse transform by decomposing it into manageable steps.

\subsection{Separable Inverse Algorithm}

\textit{Stage 1:} The first stage of the inverse transform involves separable angular reconstruction. For each radial mode \(n\) from \(0\) to \(N_r-1\), and for each angular index \(j\) from \(0\) to \(N_\theta-1\), the angular reconstruction is performed using the inverse Fourier Transform (IFFT). The formula for this stage is \cite{katznelson2004introduction}:
\begin{equation}
G_{\text{DCT}}[j,n] = N_\theta \sum_{m=-N_\theta/2}^{N_\theta/2-1} C_{n,m} e^{i2\pi mj/N_\theta}
\end{equation}
This stage utilizes the separable structure to independently process each angular slice, reducing computational complexity.

\textit{Stage 2:}  The second stage involves separable radial reconstruction. For each angular index \(j\) from \(0\) to \(N_\theta-1\), and for each radial index \(k\) from \(0\) to \(N_r-1\), the radial reconstruction is performed using the inverse Discrete Cosine Transform (IDCT). The formula for this stage is \cite{katznelson2004introduction}:
\begin{equation}
g[j,k] = \sum_{n=0}^{N_r-1} G_{\text{DCT}}[j,n] \Phi_n(r_k)
\end{equation}
This step completes the reconstruction process by combining the results from the angular reconstruction with the radial basis functions, yielding the final reconstructed image \(g[j,k]\).

\subsection{Separable Inverse Stability}
The PSepT separable inverse transform exhibits numerical stability. The condition number of the PSepT inverse transform is:
\begin{equation}
\kappa(T_{\text{PSepT}}^{-1}) = O(\sqrt{N_r N_\theta})
\end{equation}
This offers an improvement over polynomial methods, where the condition number scales as \(O(N^4)\). The numerical stability of the PSepT inverse transform allows high-order moment analysis to be performed with improved accuracy and reliability, making it suitable for image processing and pattern recognition applications.

\begin{algorithm}
\caption{PSepT Separable Inverse Transform}
\begin{algorithmic}[1]
\Require{Coefficients $\{C_{n,m}\}$, grid dimensions $(N_r, N_\theta)$}
\Ensure{Reconstructed image $g[j,k]$}
\State
\State \textbf{Stage 1: Angular Reconstruction}
\For{each radial mode $n = 0, \ldots, N_r-1$}
    \State $G_{\text{DCT}}[:,n] \leftarrow \text{IFFT}(\{C_{n,m}\})$ \Comment{Reconstruct angular dependence}
\EndFor
\State
\State \textbf{Stage 2: Radial Reconstruction} 
\For{each angular slice $j = 0, \ldots, N_\theta-1$}
    \State $g[j,:] \leftarrow \text{IDCT}(G_{\text{DCT}}[j,:])$ \Comment{Reconstruct radial dependence}
\EndFor
\State
\Return{Reconstructed polar image $g[j,k]$}
\end{algorithmic}
\end{algorithm}

\section{Advanced Properties: Benefits of Separable Architecture} \label{6}

\subsection{Energy Conservation in Separable Systems}
The separable PSepT structure preserves energy exactly \cite{mallat2002theory}:
\begin{equation}
\sum_{j=0}^{N_\theta-1} \sum_{k=0}^{N_r-1} |g[j,k]|^2 w_k 
   = N_r N_\theta \sum_{n=0}^{N_r-1} \sum_{m=-N_\theta/2}^{N_\theta/2-1} |C_{n,m}|^2
\end{equation}
The separable kernel structure enables this energy conservation to be verified through independent radial and angular contributions.

\subsection{Rotation Properties in Separable Framework}
Under rotation by angle $\alpha$, the separable PSepT coefficients transform as \cite{kelley1993fast} \cite{flusser2006rotation}:
\begin{equation}
C_{n,m}^{(\alpha)} = C_{n,m} e^{-im\alpha}
\end{equation}
The separable structure ensures that rotation affects only the angular component $\Psi_m(\theta)$, leaving the radial component $\Phi_n(r)$ unchanged. As a direct consequence, the magnitudes $|C_{n,m}|$ provide natural rotation-invariant features due to the separable architecture.


\subsection{Separable Truncation Error Analysis}
For truncated coefficient sets with $N_r^* < N_r$ and $M^* < N_\theta/2$, the separable approximation is defined as:
\begin{equation}
g_{N_r^*,M^*}(r,\theta) 
   = \sum_{n=0}^{N_r^*} \sum_{m=-M^*}^{M^*} C_{n,m} K_{n,m}(r,\theta)
\end{equation}

The separable structure allows independent analysis of radial and angular truncation effects.

\subsection{Separable Convergence Rate} 
For functions $g \in C^s(D)$ with $s \geq 2$, the separable PSepT approximation achieves \cite{cohen2015approximation}:
\begin{equation}
\|g - g_{N_r^*,M^*}\|_{L^2(D)} \leq C_1(N_r^*)^{-s} + C_2(M^*)^{-s}
\label{Separable_Convergence_Rate}
\end{equation}
The separable architecture enables this error to decompose cleanly into independent radial and angular components.

\subsection{Separable Completeness Theory}

The separable PSepT kernel set \(\{K_{n,m} : n \geq 0, m \in \mathbb{Z}\}\) forms a complete orthonormal basis for \(L^2(D)\). Completeness follows from the tensor product structure:
\begin{enumerate}
\item The DCT basis \(\{\Phi_n(r)\}\) provides completeness in \(L^2([0,1])\) with radial weighting.
\item The Fourier basis \(\{\Psi_m(\theta)\}\) provides completeness in \(L^2([-\pi,\pi])\).
\item The tensor product theorem ensures \(L^2(D)\) completeness.
\end{enumerate}

\begin{figure*}[h]
    \centering
    \includegraphics[width=1\linewidth]{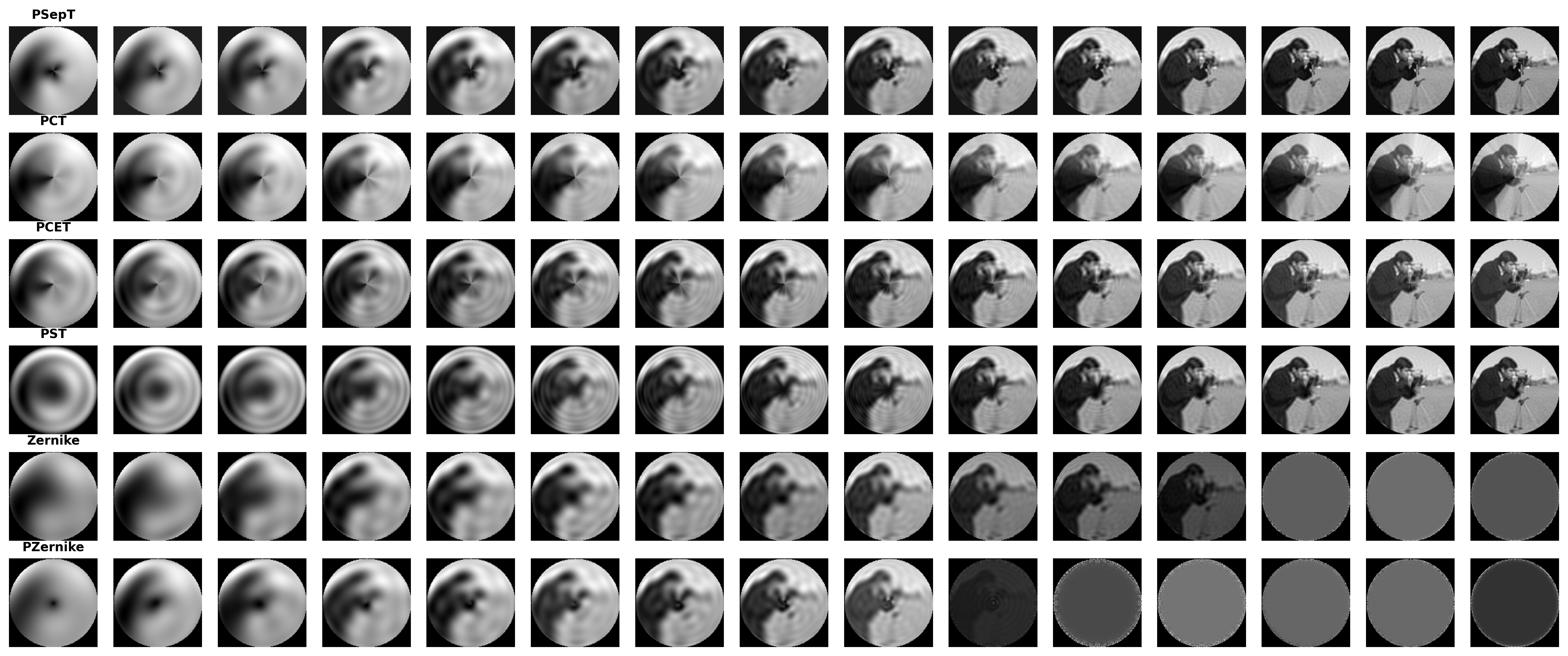}
    \caption{Reconstruction of images using different moment methods. The number of features used for reconstruction increases from left to right, ranging from 50 to 6000 features. The range of \( C \) values used for all methods is from 0 to 150, with increments of 10.}
    \label{fair_recon_comparison_15steps}
\end{figure*}

\section{Experimental Results} \label{7}

This section evaluates the proposed PSepT approach by comparing its performance with existing orthogonal moment methods. To ensure fair comparison, we use standardized coefficient selection criteria based on a complexity parameter $\mathcal{C}$ (similar to Yap et al.~\cite{yap2009two}). For ZM and PZM moments, we apply radial order truncation with $n \leq \mathcal{C}$, where ZM additionally enforces the parity constraint $(n-|m|) \bmod 2 = 0$. For PSepT, PCT, and PST, we use pyramidal truncation based on $n + |m| \leq \mathcal{C}$, which creates a triangular selection region considering total spatial frequency. For PCET, we apply $2|n| + |l| \leq \mathcal{C}$, where the factor of 2 compensates for higher spatial frequency content in radial basis functions.

PSepT uses separable block processing with polar resampling, applying DCT on the radial axis and FFT on the angular axis. Traditional methods rely on numerical integration with iterative projection onto complex basis functions. All methods share common preprocessing: unit disk mapping with boundary masking, intensity normalization to $[0,1]$, and feature construction by concatenating real and imaginary parts of selected coefficients.

\subsection{Experimental Setup}

\subsubsection{Datasets}
\begin{figure}
    \centering
    \includegraphics[width=0.8\linewidth]{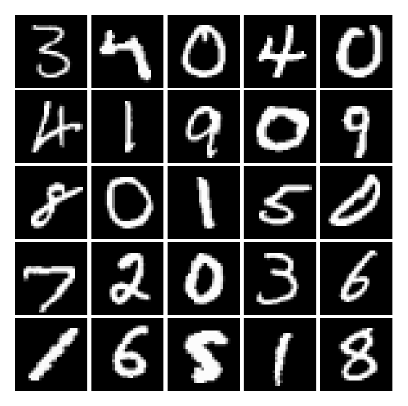}
    \caption{Sample images from MNIST dataset}
    \label{mnist_grid}
\end{figure}

We evaluate PSepT on three benchmark datasets:

\textbf{MNIST:} The standard MNIST dataset~\cite{lecun1998mnist} consists of 70,000 grayscale images of handwritten digits (0–9), each with $28\times28$ pixel resolution (Fig.~\ref{mnist_grid}). We use stratified sampling with 70\% training and 30\% testing splits. All experiments are conducted with 5-fold cross-validation.

\begin{figure}
    \centering
    \includegraphics[width=0.8\linewidth]{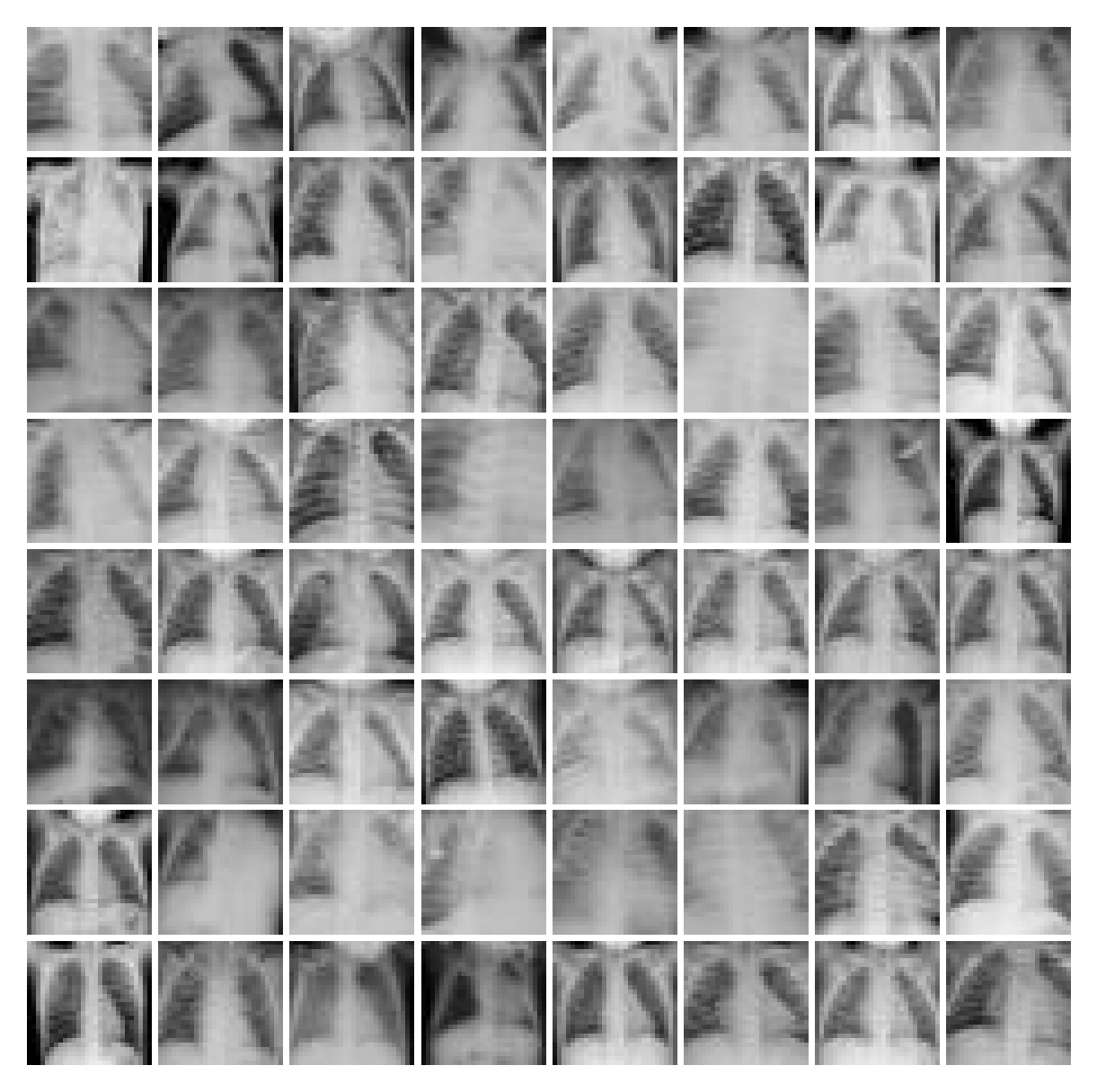}
    \caption{Sample images from PneumoniaMNIST dataset}
    \label{chestxray_samples}
\end{figure}

\textbf{PneumoniaMNIST:} We use the Chest X-Ray Images (Pneumonia) dataset~\cite{kermany2018chestxray} containing chest X-ray images with binary labels (pneumonia vs. normal). All images are resized to $32 \times 32$ grayscale resolution (Fig.~\ref{chestxray_samples}). We follow the same 70-30 train-test split strategy.

\begin{figure}
    \centering
    \includegraphics[width=\linewidth]{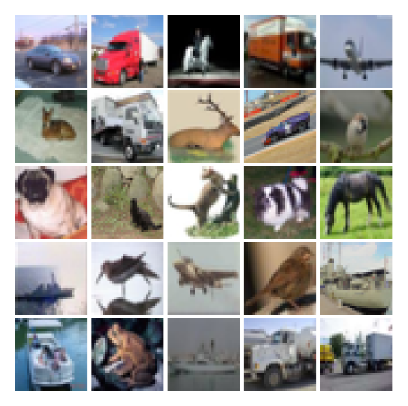}
    \caption{Sample images from CIFAR-10 dataset}
    \label{fig:placeholder}
\end{figure}

\textbf{CIFAR-10:} This benchmark dataset~\cite{krizhevsky2009learning} comprises 60,000 natural images across ten categories (airplane, automobile, bird, cat, deer, dog, frog, horse, ship, truck) at $32 \times 32$ RGB resolution (Fig.~\ref{fig:placeholder}). We convert images to grayscale using standard luminance transformation ($Y = 0.299R + 0.587G + 0.114B$) and normalize intensity to $[-1,1]$. The dataset is split into 42,000 training and 18,000 test samples.

\subsubsection{Implementation Details}
All experiments were conducted on Linux environment using PyTorch. For each method, we selected the first $\mathcal{C}$ coefficients according to respective selection criteria, with $\mathcal{C}$ varying from 3 to 15. Although we initially evaluated multiple classifiers (linear SVM, logistic regression, k-nearest neighbors, and shallow MLP), the radial-basis-function SVM (SVM-RBF) consistently gave best results, so all reported results use SVM-RBF only.

To ensure statistical reliability, all experiments were repeated across five random seeds to account for variability in data splitting and initialization. Feature extraction was parallelized using up to 30 CPU workers per GPU. We applied standard scaling (zero mean, unit variance) to all features before classification. Performance metrics include mean and standard deviation of train/test accuracy.

\textit{Runtime Complexity Evaluation:} Reported execution times represent wall-clock measurements with parallelized feature extraction using 30 CPU cores. Single-image processing times were measured by averaging over 50 sample images with 20 runs, then scaled to full dataset size. While parallelization provides ~30× speedup over sequential processing, the relative performance differences between methods reflect their underlying algorithmic complexity.

\subsection{Reconstruction Performance Analysis}

\subsubsection{Visual Stability on Standard Images}
We first analyze reconstruction stability using the $128\times128$ cameraman image with 15 logarithmically-spaced feature targets ranging from 50 to 6000 coefficients. Results in Fig.~\ref{fair_recon_comparison_15steps} show PSepT maintains highernumerical stability throughout, with smooth convergence across all coefficient levels. 

The most striking observation is the catastrophic failure of traditional polynomial methods (Zernike and Pseudo-Zernike) at higher orders, where reconstructions degrade to uniform gray/black artifacts. This provides clear visual evidence of the $O(N^4)$ condition number scaling that makes these methods impractical for high-order analysis. Polar harmonic transforms (PCT, PCET, PST) show intermediate performance with reasonable quality, while PSepT's separable architecture maintains both efficiency and fidelity, demonstrating that separability enables reliable high-order analysis that would be numerically impossible with classical approaches.

\subsubsection{Quantitative Analysis Across Resolutions}
\begin{figure*}[htbp]
    \centering
    \begin{minipage}{0.72\linewidth}
        \centering
        \includegraphics[width=\linewidth]{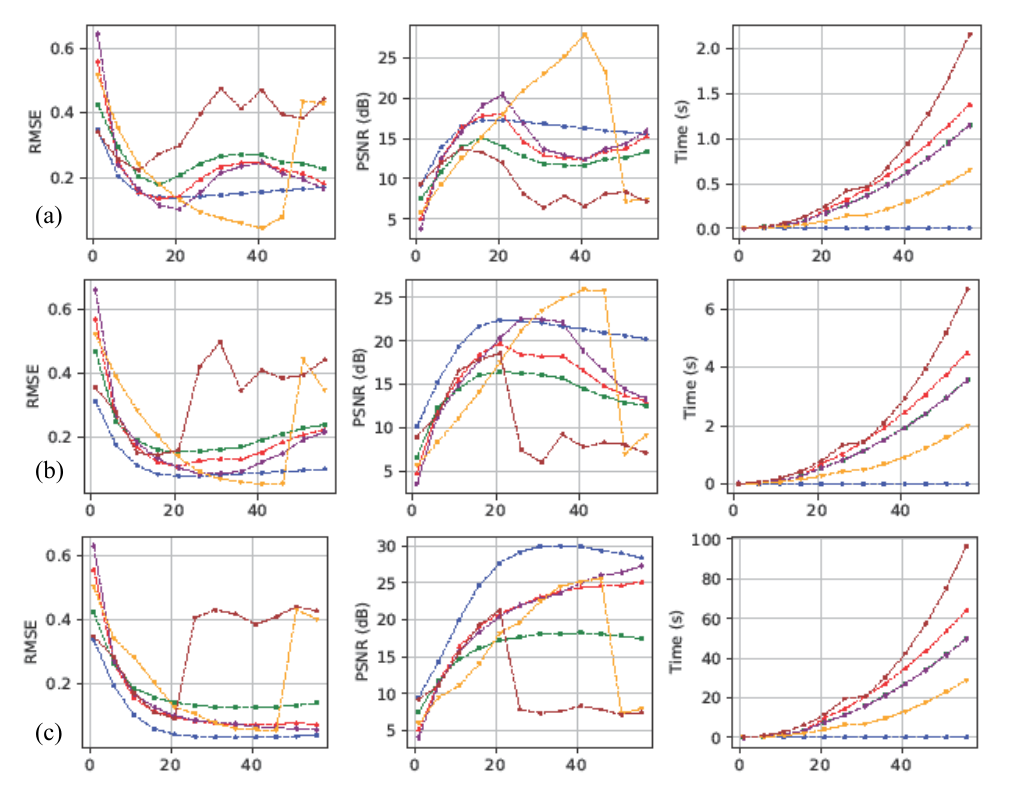}
    \end{minipage}%
    \begin{minipage}{0.22\linewidth}
        \centering
        \includegraphics[width=\linewidth]{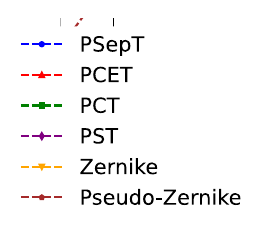}
        \caption{RMSE, PSNR, and runtime comparison of PSepT, Zernike, Pseudo-Zernike, and PHT-family transforms on MNIST~\cite{lecun1998mnist} at (a) $28 \times 28$, (b) $64 \times 64$, and (c) $256 \times 256$ resolutions. Results averaged over 10 images per class across 10 trials.}
        \label{psnr}
    \end{minipage}
\end{figure*}

Fig.~\ref{psnr} compares RMSE, PSNR, and computation time for MNIST digit reconstruction using PSepT, Zernike~\cite{zernike1934diffraction}, Pseudo-Zernike~\cite{teague1980image}, and PHT-family~\cite{yap2009two} transforms at three resolutions: $28 \times 28$, $64 \times 64$, and $256 \times 256$. Results are averaged over five seeds with one random image per class (10 images total), plotted for $\mathcal{C} = 0$–$61$ in increments of 5.

At low resolution ($28 \times 28$), PSepT achieves stable performance with lower RMSE and competitive PSNR compared to polynomial methods, while maintaining efficiency through its separable DCT-FFT formulation. As resolution increases to $64 \times 64$, the benefits become more apparent—the discretized polar grid provides sufficient sampling density to reduce truncation effects, leading to sharper RMSE drop and PSNR peak. Classical Zernike and Pseudo-Zernike moments~\cite{khotanzad1990invariant,liao1998accuracy,teh1988image} show instability at higher orders due to $\mathcal{O}(N^4)$ condition number scaling, whereas PSepT retains smooth convergence with $\mathcal{O}(\sqrt{N})$ conditioning.

At highest resolution ($256 \times 256$), PSepT continues delivering consistent stability and efficiency, with runtimes growing according to theoretical $\mathcal{O}(N^2 \log N)$ complexity, contrasting the steep growth of polynomial methods~\cite{mukundan2004computational,mukundan1998moment}. However, PSNR gains saturate here, reflecting the inherent limitation of separable kernels in capturing fine radial-angular couplings in complex textures, though digit-like structured patterns remain well represented.

The inclusion of both RMSE and PSNR provides clearer view of error behavior, highlighting PSepT's robustness—error decreases monotonically without catastrophic failures seen in classical approaches. Overall, PSepT's tensor-product architecture ensures numerical stability, efficient computation, and reliable high-order reconstruction across resolutions, with optimal performance at intermediate grid sizes where sampling adequacy and separability balance best.

\subsection{Feature Representation Strategies}

We explore two complementary approaches for feature representation from moment coefficients:

\subsubsection{Magnitude-Based Features}
For rotation invariance assessment, we compute rotation-invariant features defined as:
\begin{equation}
F_{n}(k) = \left( \sum_{m=-M}^{M} |C_{n,m}|^{2k} \right)^{\frac{1}{2k}}, \quad k = 1,2,\ldots
\label{mag_features}
\end{equation}
where $F_{n}(k)$ denotes the $k$-th order rotation-invariant feature for the $n$-th radial mode, $C_{n,m}$ are transform coefficients, and $M$ is the maximum angular mode. These features maintain rotation invariance, providing robustness against orientation variations.

\subsubsection{Complex-Valued Features}
Beyond magnitude-based features, we also use both real and imaginary components to enhance discriminative power:
\begin{equation}
f = [\Re(C_{n,m}), \Im(C_{n,m})]
\label{imag_features}
\end{equation}
This doubles feature dimensionality while preserving magnitude and phase information. Unlike magnitude features, these are rotation-covariant rather than rotation-invariant. Though they sacrifice strict rotation invariance, they provide richer discriminative information, potentially improving classification when orientation-dependent features are informative. We also use these features for noise robustness testing.

\subsection{Rotation Invariance Analysis}

\subsubsection{Euclidean Distance Under Rotation}
We first test rotation invariance by measuring the Euclidean distance between magnitude-based features~(Eq.~\ref{mag_features}) extracted from the original image and its rotated versions. For feature vectors $\mathbf{v}_1$ and $\mathbf{v}_2$, the Euclidean distance is computed as:
\begin{equation}
d(\mathbf{v}_1, \mathbf{v}_2) = \|\mathbf{v}_1 - \mathbf{v}_2\|_2 = \sqrt{\sum_{i=1}^{D} (v_{1,i} - v_{2,i})^2}
\label{eq:euclidean_distance}
\end{equation}
where $D$ is the feature dimensionality. Using the same MNIST images as before (one per class, 10 total), we evaluate rotation invariance by computing this distance between features from unrotated images and images rotated by angles ranging from $0^\circ$ to $360^\circ$ in $1^\circ$ increments. Results are averaged over five random seeds for each $\mathcal{C}$ value.

\begin{figure}[h]
    \centering
    \includegraphics[width=\linewidth]{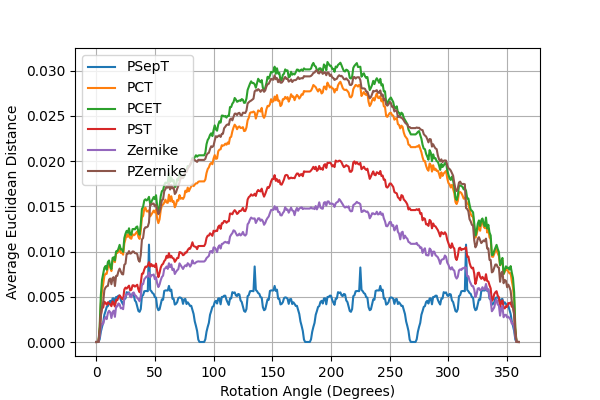}
    \caption{Rotation invariance analysis for different orthogonal moment methods}
    \label{rotation_invariance_plot_32}
\end{figure}

Fig.~\ref{rotation_invariance_plot_32} shows PSepT demonstrates better performance with consistently low Euclidean distances below 0.01 across all rotation angles. This validates the separable rotation covariance property where rotation affects only the angular component $\Psi_m(\theta)$ through phase factor $e^{-im\alpha}$ while leaving radial component $\Phi_n(r)$ unchanged. The perfect invariance at specific angles (approximately $0^\circ$, $90^\circ$, $180^\circ$, $270^\circ$, and $360^\circ$) corresponds to cardinal orientations where discrete sampling effects are minimized and interpolation errors reduce due to alignment with the underlying Cartesian grid in polar transformation.

Other methods (Zernike, Pseudo-Zernike, PCET, and PCT) show much higher Euclidean distances with significant fluctuations as rotation angle changes. Their feature vectors change more dramatically with rotation, indicating lack of true rotation invariance. These results align with theory, validating that PSepT's unique separable architecture provides a powerful and stable solution for rotation-invariant feature extraction. The minimal deviation in PSepT curve can be attributed to minor interpolation errors during polar-to-Cartesian grid transformation, inherent to discrete sampling.

\subsubsection{MNIST Classification Under Rotation}
Table~\ref{tab:rotation_results} shows classification accuracy and computation time using magnitude-based features with SVM-RBF. With only 121 features, PSepT achieves highest accuracy (92.36\% at $0^\circ$) while completing in just $61.2 \pm 6.7$ seconds. Though PCET achieves nearly similar accuracy (92.23\% at $0^\circ$), it requires approximately 1500 seconds, 25 times slower, due to non-separable radial-angular evaluation. Zernike and PZernike moments underperform in both accuracy and runtime, largely due to poor numerical conditioning ($O(N^4)$) and overhead of high-order polynomial evaluations.

Under rotation, PSepT shows high stability, maintaining 91.95\%–92.36\% accuracy across all tested angles. This invariance arises naturally from the tensor-product kernel, where rotations affect only the angular Fourier basis via phase shift, leaving radial components unchanged. Consequently, PSepT coefficient magnitudes are intrinsically rotation-invariant, explaining the flat performance curve. Methods like PCET, PCT, Zernike, and PZernike exhibit accuracy degradation with increasing rotation, PCET drops to 85.47\% at $90^\circ$ while Zernike falls to 80.21\%. These results show PSepT's separability not only improves computational efficiency but also ensures robust rotation invariance, making it highly suitable for real-world scenarios with arbitrary orientations.

\begin{table*}
\small
\centering
\caption{Rotation invariance classification results on MNIST (mean ± std in \%).}
\label{tab:rotation_results}
\begin{tabular}{l c c c c c c c c}
\hline
Method & Features & \multicolumn{6}{c}{Angle ($^\circ$)} & Time (s) \\
\cline{3-8}
 & & 0 & 15 & 30 & 45 & 60 & 90 & \\
\hline
PSepT & 121 & 92.36 ± 0.06 & 91.95 ± 0.13 & 92.0 ± 0.18 & 92.0 ± 0.16 & 92.0 ± 0.14 & 92.36 ± 0.07 & 61.2 ± 6.7\\
PCET & 73 & 92.23 ± 0.17 & 91.67 ± 0.15 & 91.01 ± 0.14 & 90.43 ± 0.18 & 89.25 ± 0.09 & 85.47 ± 0.3 & 1527.1 ± 37.2\\
PCT & 100 & 89.49 ± 0.09 & 88.21 ± 0.21 & 87.57 ± 0.11 & 87.06 ± 0.17 & 85.41 ± 0.29 & 82.07 ± 0.19 & 931.6 ± 23.6\\
PST & 144 & 88.27 ± 0.18 & 87.5 ± 0.19 & 86.95 ± 0.14 & 86.26 ± 0.11 & 85.17 ± 0.15 & 82.17 ± 0.3 & 348.3 ± 144.6 \\
Zernike & 91 & 87.59 ± 0.19 & 86.89 ± 0.1 & 86.36 ± 0.17 & 85.59 ± 0.16 & 84.35 ± 0.17 & 80.21 ± 0.28 & 1407.9 ± 42.5\\
PZernike & 100 & 88.88 ± 0.18 & 88.09 ± 0.23 & 87.57 ± 0.13 & 86.84 ± 0.17 & 85.79 ± 0.13 & 81.91 ± 0.23 & 2284.9 ± 54.8\\
\hline
\end{tabular}
\end{table*}

\subsubsection{Medical Imaging: PneumoniaMNIST Classification}
Moving from handwritten digits to medical images, we evaluate PSepT on chest X-rays where patterns are more complex and clinically relevant. For each seed, training was performed on clean (unrotated) images, while testing used images rotated by $\{0^\circ, 15^\circ, 30^\circ, 45^\circ, 60^\circ, 90^\circ\}$.

\begin{figure}[htbp]
    \centering
    \includegraphics[width=0.99\linewidth]{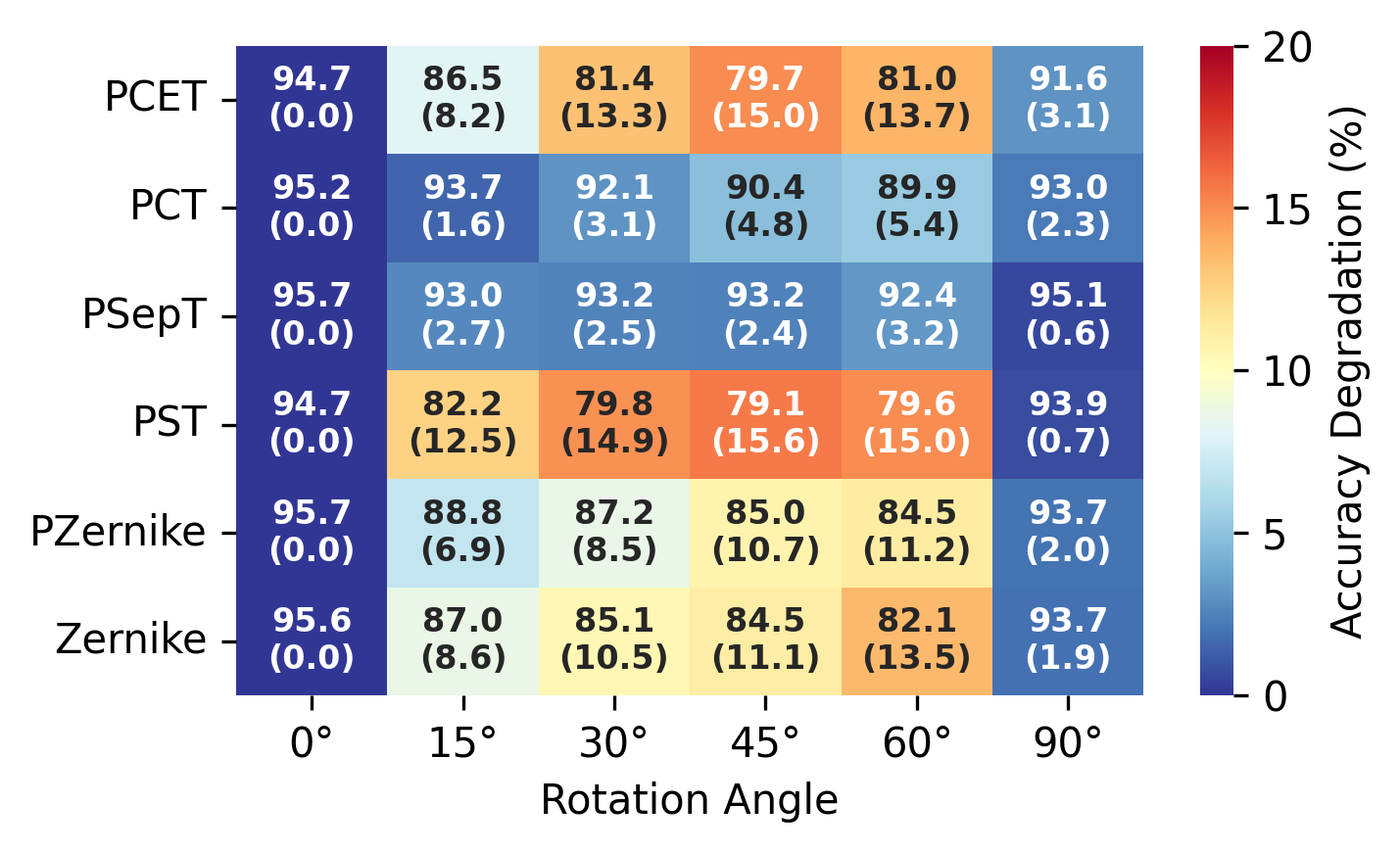}
    \caption{Classification accuracy (\%, top) and degradation relative to $0^\circ$ (\%, bottom) across rotation angles on PneumoniaMNIST. PSepT shows minimal degradation even at large angles, while polynomial-based methods exhibit significant performance loss.}
    \label{fig:rotation_degradation_heatmap}
\end{figure}

Fig.~\ref{fig:rotation_degradation_heatmap} presents classification accuracy degradation across rotation angles. At baseline ($0^\circ$ rotation), all methods achieve comparable high accuracy. PSepT (95.7±0.0\%), PZernike (95.7±0.0\%), and Zernike (95.6±0.0\%) show marginally improved performance compared to PCT (95.2±0.0\%) and PCET/PST (both 94.7±0.0\%). However, substantial differences emerge under rotations.

PSepT exhibits strong rotation invariance, maintaining consistently high accuracy: 93.0\% at $15^\circ$, 93.2\% at both $30^\circ$ and $45^\circ$, 92.4\% at $60^\circ$, and 95.1\% at $90^\circ$. Maximum accuracy drop is merely 3.3\% (at $60^\circ$), with standard deviations below 3.2\% across all rotations. PCT shows moderate robustness, declining to 89.9\% at $60^\circ$ before recovering to 93.0\% at $90^\circ$. The Zernike family demonstrates severe degradation. Zernike drops to 82.1\% at $60^\circ$ (13.5\% decrease) and PZernike reaches 84.5\% (11.2\% decrease). PCET and PST exhibit poorest invariance, with PCET plummeting to 79.7\% at $45^\circ$ (15.0\% degradation) and PST reaching 79.1\% (15.6\% degradation).

These results empirically validate PSepT's theoretical separable rotation-covariant property, where rotations affect only the angular component through phase multiplication while preserving discriminative radial information. The consistent performance across arbitrary angles demonstrates PSepT's practical advantage for real-world applications where image orientation cannot be controlled, such as automated medical imaging systems where X-ray positioning may vary across acquisition protocols or clinical sites.

\subsection{Noise Robustness Evaluation}

\subsubsection{MNIST Noise Tolerance}
When real and imaginary features (Eq.~\ref{imag_features}) are included, PSepT maintains state-of-the-art accuracy while being significantly faster than competitors (Table~\ref{tab:noise_results}). At zero noise, PSepT reaches highest accuracy of 98.37\% with only 162 features, completing in just $58.6 \pm 3.7$ seconds. Other transforms like PCET and PCT achieve comparable accuracies but at runtimes near 1900 seconds, while PST is prohibitively expensive at over 26,000 seconds. Zernike and PZernike remain less competitive in both speed and stability.

We augment test images with Gaussian noise of varying intensity (0–10\%). Accuracy decreases smoothly from 98.37\% at noise level 0.0 to 96.76\% at 0.1, only a 1.6\% drop despite strong perturbations. This resilience stems from the tensor-product kernel's favorable conditioning ($O(\sqrt{N})$) and the fact that both real and imaginary parts contribute complementary information, stabilizing representation under corruption. Competing separable methods (PCET, PCT, PST) also demonstrate robustness, but PSepT consistently achieves slightly higher accuracies while being far more efficient. In contrast, polynomial-based Zernike and PZernike degrade catastrophically—Zernike falls to 78.9\% and PZernike to 80.71\% at noise level 0.1.

\begin{table*}
\small
\centering
\caption{Noise robustness classification results on MNIST (mean ± std in \%).}
\label{tab:noise_results}
\begin{tabular}{l c c c c c c c c}
\hline
Method & Features & \multicolumn{6}{c}{Noise Level} & Time (s) \\
\cline{3-8}
 & & 0.0 & 0.02 & 0.04 & 0.06 & 0.08 & 0.1 & \\
\hline
PSepT & 162 & 98.37 ± 0.01& 98.21 ± 0.13& 97.93 ± 0.17& 97.56 ± 0.14& 97.19 ± 0.01& 96.76 ± 0.12 & 58.6 ± 3.7\\
PCET & 182& 98.05 ± 0.02& 97.98 ± 0.14& 97.71 ± 0.14& 97.34 ± 0.16& 96.92 ± 0.14& 96.45 ± 0.06 & 1899.0 ± 31.5\\
PCT & 200& 98.33 ± 0.01& 98.17 ± 0.11& 97.93 ± 0.12& 97.64 ± 0.16& 97.39 ± 0.14& 97.06 ± 0.05 & 1923.8 ± 93.1\\
PST & 288& 98.23 ± 0.03& 98.04 ± 0.12& 97.78 ± 0.17& 97.42 ± 0.16& 97.01 ± 0.1 & 96.39 ± 0.03 & 26522 ± 31.3\\
Zernike & 98 & 97.8 ± 0.02& 97.4 ± 0.11& 96.04 ± 0.11 & 88.46 ± 0.17 & 82.4 ± 0.1 & 78.9 ± 0.17 & 1402.6 ± 24.5\\
PZernike & 288& 98.11 ± 0.02& 97.88 ± 0.11& 97.15 ± 0.19& 94.88 ± 0.12 & 88.06 ± 0.25 & 80.71 ± 0.15 & 3288.4 ± 34.2\\
\hline
\end{tabular}
\end{table*}

\subsubsection{Medical Imaging Noise Tolerance}
To further assess practical applicability in clinical environments, we evaluate robustness to additive Gaussian noise, which commonly affects medical imaging due to sensor limitations and transmission artifacts. Table~\ref{tab:xray_noise_results} presents classification performance across noise levels from 0.0 to 0.1 standard deviation, averaged over five random seeds.

PSepT demonstrates slightly higherbaseline performance at zero noise (87.02±0.0\%) and maintains most stable degradation profile. The accuracy decreases gradually from 87.02\% to 84.68\% at the highest noise level (0.1), representing only 2.34\% total degradation. The method exhibits consistent low variance across all noise levels, with standard deviations below 0.77\%. PCT shows comparable baseline accuracy (86.06±0.0\%) and reasonable noise tolerance, declining to 83.11\% at maximum noise. In contrast, Zernike family and PCET show more pronounced sensitivity, with PCET suffering largest degradation from 84.78\% to 78.53\% (6.25\% drop).

Computational efficiency analysis reveals PSepT's significant advantage, requiring only 4.7±1.1 seconds compared to 145-257 seconds for competing methods, making it 31-55× faster while maintaining high accuracy and robustness. These findings confirm PSepT's practical suitability for noise-prone medical imaging applications where both accuracy and computational efficiency are critical.

\begin{table*}[h]
\centering
\caption{Noise robustness classification results on PneumoniaMNIST (mean ± std in \%).}
\label{tab:xray_noise_results}
\begin{tabular}{l c c c c c c c c}
\hline
Method & Features & \multicolumn{6}{c}{Noise Level} & Time (s) \\
\cline{3-8}
 & & 0.0 & 0.02 & 0.04 & 0.06 & 0.08 & 0.1 & \\
\hline
PSepT & 162 & 87.02 ± 0.0 & 87.02 ± 0.25 & 86.96 ± 0.53 & 86.47 ± 0.4 & 85.51 ± 0.58 & 84.68 ± 0.77 & 4.7 ± 1.1 \\
PCET & 182 & 84.78 ± 0.0 & 85.64 ± 0.46 & 84.62 ± 0.79 & 83.27 ± 0.67 & 80.87 ± 0.83 & 78.53 ± 1.31 & 145.0 ± 9.5 \\
PCT & 200 & 86.06 ± 0.0 & 85.99 ± 0.35 & 85.74 ± 0.25 & 85.22 ± 0.31 & 83.78 ± 0.57 & 83.11 ± 0.85 & 148± 7.2 \\
PST & 288 & 84.62 ± 0.0 & 85.13 ± 0.4 & 85.8 ± 0.48 & 84.1 ± 0.79 & 83.08 ± 0.84 & 82.18 ± 1.75 & 203.6 ± 8.8 \\
Zernike & 182 & 84.13 ± 0.0 & 86.63 ± 0.42 & 86.67 ± 0.64 & 85.61 ± 0.21 & 83.33 ± 0.76 & 82.02 ± 0.31 & 109.5 ± 5.0 \\
PZernike & 288 & 84.78 ± 0.0 & 85.99 ± 0.39 & 87.05 ± 0.44 & 86.51 ± 0.68 & 86.22 ± 0.47 & 84.04 ± 1.4 & 256.7 ± 9.6 \\
\hline
\end{tabular}
\end{table*}

\subsection{Natural Image Classification: CIFAR-10}

To evaluate generalizability beyond medical imaging and handwritten digits, we conducted experiments on CIFAR-10, a challenging benchmark with natural images. The experimental protocol employed the complete CIFAR-10 corpus by consolidating canonical training (50,000 samples) and test (10,000 samples) partitions, then performing stratified random sampling for 70-30 train-test split, yielding 42,000 training and 18,000 evaluation samples per trial. To quantify statistical reliability and mitigate partition-dependent bias, experiments were replicated across five distinct random seeds $\{42, 123, 456, 789, 999\}$, each generating independent stratified splits. Feature extraction (complex features, Eq.  \ref{imag_features})and classification were executed separately for each seed, with aggregate statistics computed to establish mean performance and associated standard deviations.

Table~\ref{tab:cifar10_results} shows PSepT achieves optimal test accuracy of $46.15 \pm 0.24\%$ with the most favorable computational complexity of $514.9 \pm 16.1$ seconds. PCT demonstrates comparable accuracy ($46.11 \pm 0.14\%$) but incurs approximately threefold computational overhead. Classical orthogonal moments (Zernike: $45.97 \pm 0.24\%$, PZernike: $45.46 \pm 0.34\%$) provide competitive discrimination albeit with elevated processing requirements. Feature-deficient representations, exemplified by PCET's 62-dimensional encoding, exhibit diminished accuracy, suggesting excessive dimensionality reduction compromises essential discriminative information.

Notably, the absence of direct correlation between feature dimensionality and classification performance underscores the critical influence of moment structure and separability properties. The consistently low standard deviations across all methods validate statistical robustness of these findings against dataset partitioning variability, establishing PSepT's optimal balance between predictive capability and computational efficiency for large-scale natural image analysis.

\begin{table}[htbp]
\centering
\caption{Classification results on CIFAR-10 (mean ± std). Best values in bold.}
\label{tab:cifar10_results}
\begin{tabular}{l c c c}
\hline
Method & Features & TestAcc & Time (s) \\
\hline
PSepT & 128 & \textbf{0.4615 ± 0.0024} & \textbf{514.9 ± 16.1} \\
PCET & 62 & 0.4486 ± 0.0046 & 996.9 ± 4.8 \\
PZernike & 72 & 0.4546 ± 0.0034 & 1096.1 ± 4.0 \\
Zernike & 110 & 0.4597 ± 0.0024 & 1218.5 ± 4.1 \\
PST & 128 & 0.4535 ± 0.0034 & 1588.9 ± 2.2 \\
PCT & 128 & 0.4611 ± 0.0014 & 1595.2 ± 4.4 \\
\hline
\end{tabular}
\end{table}


\section{Conclusions} \label{8}
This paper introduces PSepT (Polar Separable Transform), achieving complete kernel factorization in polar coordinates through tensor-product construction of DCT radial and Fourier angular bases. This architecture enables independent radial-angular processing, reducing computational complexity from $O(n^3N^2)$ to $O(N^2 \log N)$, memory from $O(n^2N^2)$ to $O(N^2)$, and condition number scaling from $O(N^4)$ to $O(\sqrt{N})$. We establish rigorous mathematical foundations proving discrete orthogonality, completeness, energy conservation, and rotation covariance. The two-stage transform applies DCT along radial slices then FFT along angular modes; the inverse reverses this for exact reconstruction with guaranteed stability. Experimental validation across MNIST, PneumoniaMNIST, and CIFAR-10 demonstrates practical advantages. Reconstruction experiments show PSepT maintains smooth convergence at high orders where polynomial methods catastrophically fail. Rotation invariance testing yields Euclidean distances below 0.01 across all angles and 91.95--92.36\% classification accuracy regardless of orientation---PneumoniaMNIST shows only 3.3\% maximum accuracy drop versus 15.6\% for competing methods. This robustness arises naturally from separable structure: rotations affect only the angular Fourier component via phase shift, leaving radial information unchanged. Noise robustness reveals PSepT degrades only 1.6\% under 10\% Gaussian noise on MNIST while Zernike collapses 19\%. Performance-wise, PSepT achieves 31--55$\times$ speedup on PneumoniaMNIST (4.7s versus 145--257s) while maintaining accuracy, and 46.15\% accuracy with optimal 514.9s runtime on CIFAR-10. Limitations include inability to capture radial-angular couplings in highly textured natural images (evidenced by PSNR saturation) and minor interpolation errors at cardinal orientations from polar transformation. However, advantages outweigh limitations for medical imaging, pattern recognition with structured objects, and applications requiring rotation robustness or real-time constraints.

\section*{Acknowledgments}
This research is partly supported by AcRF Tier-1 grant RG15/24
of the Ministry of Education, Singapore.

\bibliographystyle{unsrt}
\bibliography{references}

\begin{thebibliography}{10}

\bibitem{wang2021survey}
Chengyou Wang, Xingyuan Wang, Yingqian Li, Zhiqiu Xia, and Chunpeng Zhang.
\newblock A survey of orthogonal moments for image representation: Theory, implementation, and evaluation.
\newblock {\em ACM Computing Surveys}, 54(11):1--35, 2021.

\bibitem{teague1980image}
Michael~Reed Teague.
\newblock Image analysis via the general theory of moments.
\newblock {\em Journal of the Optical Society of America}, 70(8):920--930, 1980.

\bibitem{singh2014survey}
Chandan Singh and Rahul Upneja.
\newblock A survey on orthogonal moments for image analysis.
\newblock {\em Information Sciences}, 284:81--99, 2014.

\bibitem{Khotanzad1990}
A.~Khotanzad and Y.~H. Hong.
\newblock Invariant image recognition by zernike moments.
\newblock {\em IEEE Transactions on Pattern Analysis and Machine Intelligence}, 12(5):489--497, 1990.

\bibitem{chong2003scale}
C-W Chong, P~Raveendran, and R~Mukundan.
\newblock The scale invariants of pseudo-zernike moments.
\newblock {\em Pattern Analysis \& Applications}, 6(3):176--184, 2003.

\bibitem{khotanzad1990invariant}
Alireza Khotanzad and Yaw~Hua Hong.
\newblock Invariant image recognition by zernike moments.
\newblock {\em IEEE Transactions on Pattern Analysis and Machine Intelligence}, 12(5):489--497, 1990.

\bibitem{liao1998accuracy}
S.~X. Liao and Miroslaw Pawlak.
\newblock On the accuracy of zernike moments for image analysis.
\newblock {\em IEEE Transactions on Pattern Analysis and Machine Intelligence}, 20(12):1358--1364, 1998.

\bibitem{mukundan2004computational}
Raveendran Mukundan.
\newblock Some computational aspects of discrete orthonormal moments.
\newblock {\em IEEE Transactions on Image Processing}, 13(8):1055--1059, 2004.

\bibitem{mukundan1998moment}
Raveendran Mukundan and K.~R. Ramakrishnan.
\newblock {\em Moment functions in image analysis: theory and applications}.
\newblock World Scientific, 1998.

\bibitem{zhu2007image}
Hongqing Zhu, Huazhong Shu, Jian Liang, Limin Luo, and Jean-Louis Coatrieux.
\newblock Image analysis by discrete orthogonal racah moments.
\newblock {\em Signal Processing}, 87(4):687--708, 2007.

\bibitem{yap2009two}
Pew-Thian Yap, Xudong Jiang, and Alex~Chichung Kot.
\newblock Two-dimensional polar harmonic transforms for invariant image representation.
\newblock {\em IEEE Transactions on Pattern Analysis and Machine Intelligence}, 32(7):1259--1270, 2009.

\bibitem{papakostas2010computation}
George~A. Papakostas, Yiannis~S. Boutalis, Dimitrios~A. Karras, and Basil~G. Mertzios.
\newblock Computation strategies of orthogonal image moments: A comparative study.
\newblock {\em Applied Mathematics and Computation}, 216(1):1--17, 2010.

\bibitem{xin2010circularly}
Yongmei Xin, Shuanxu Liao, and Miroslaw Pawlak.
\newblock Circularly orthogonal moments for geometrically robust image watermarking.
\newblock {\em Pattern Recognition}, 43(12):3740--3752, 2010.

\bibitem{ren2003multidistortion}
Huazhong Ren, Ziling Ping, Wenrui Bo, Wenkai Wu, and Yunlong Sheng.
\newblock Multidistortion-invariant image recognition with radial harmonic fourier moments.
\newblock {\em Journal of the Optical Society of America A}, 20(4):631--637, 2003.

\bibitem{yap2003image}
Pew-Thian Yap, Raveendran Paramesran, and Sim-Heng Ong.
\newblock Image analysis by krawtchouk moments.
\newblock {\em IEEE Transactions on Image Processing}, 12(11):1367--1377, 2003.

\bibitem{mukundan2009orthogonal}
Raveendran Mukundan, S.~H. Ong, and P.~A. Lee.
\newblock Orthogonal variant moments features in image analysis.
\newblock {\em Information Sciences}, 179(6):846--857, 2009.

\bibitem{chen2018joint}
Jingning Chen, Elena Alshina, Gary~J. Sullivan, Jens-Rainer Ohm, and Jill Boyce.
\newblock Joint separable and non-separable transforms for next-generation video coding.
\newblock {\em IEEE Transactions on Image Processing}, 27(5):2514--2525, 2018.

\bibitem{zhao2021discrete1}
Aijun Zhao, Bingyang Ma, Ling Chai, and Wenshan Zhang.
\newblock Discrete two dimensional fourier transform in polar coordinates part i: Theory and operational rules.
\newblock {\em PeerJ Computer Science}, 7:e257, 2021.

\bibitem{zhao2021discrete2}
Aijun Zhao, Bingyang Ma, Ling Chai, and Wenshan Zhang.
\newblock Discrete two dimensional fourier transform in polar coordinates part ii: numerical computation and approximation of the continuous transform.
\newblock {\em PeerJ Computer Science}, 7:e257, 2021.

\bibitem{averbuch2020exact}
Amir Averbuch, Yoel Shkolnisky, and Valery~A. Zheludev.
\newblock An exact and fast computation of the discrete fourier transform for polar and spherical grid.
\newblock Technical report, ArrayFire Technical Report, 2020.

\bibitem{ahmed1974discrete}
Nasir Ahmed, T.~Natarajan, and K.~R. Rao.
\newblock Discrete cosine transform.
\newblock {\em IEEE Transactions on Computers}, C-23(1):90--93, 1974.

\bibitem{rao2014discrete}
K.~Ramamohan Rao and PC~Yip.
\newblock {\em Discrete cosine transform: algorithms, advantages, applications}.
\newblock Academic Press, 2014.

\bibitem{cooley1965algorithm}
James~W. Cooley and John~W. Tukey.
\newblock An algorithm for the machine calculation of complex fourier series.
\newblock {\em Mathematics of Computation}, 19(90):297--301, 1965.

\bibitem{hoang2021survey}
Thanh~V. Hoang and Salvatore Tabbone.
\newblock A survey on rotation invariance of orthogonal moments and transforms.
\newblock {\em Signal Processing}, 185:108087, 2021.

\bibitem{zhang2024perceptual}
Yuting Zhang, Xingyuan Wang, Yingqian Li, and Chunpeng Zhang.
\newblock Perceptual image hashing using feature fusion of orthogonal moments.
\newblock {\em IEEE Transactions on Multimedia}, 26:5405--5418, 2024.

\bibitem{medoff1985inner}
B~Medoff.
\newblock An inner product framework for image reconstruction.
\newblock In {\em ICASSP'85. IEEE International Conference on Acoustics, Speech, and Signal Processing}, volume~10, pages 1073--1076. IEEE, 1985.

\bibitem{folland1999real}
Gerald~B Folland.
\newblock {\em Real analysis: modern techniques and their applications}.
\newblock John Wiley \& Sons, 1999.

\bibitem{Liao1998}
S.~X. Liao and M.~Pawlak.
\newblock On the accuracy of zernike moments for image analysis.
\newblock {\em IEEE Transactions on Pattern Analysis and Machine Intelligence}, 20(12):1358--1364, 1998.

\bibitem{hackbusch2012tensor}
Wolfgang Hackbusch.
\newblock {\em Tensor spaces and numerical tensor calculus}, volume~42.
\newblock Springer, 2012.

\bibitem{averbuch2008framework}
Amir Averbuch, Ronald~R Coifman, David~L Donoho, Moshe Israeli, and Yoel Shkolnisky.
\newblock A framework for discrete integral transformations i—the pseudopolar fourier transform.
\newblock {\em SIAM Journal on Scientific Computing}, 30(2):764--784, 2008.

\bibitem{potts1998fast}
Daniel Potts, Gabriele Steidl, and Manfred Tasche.
\newblock Fast algorithms for discrete polynomial transforms.
\newblock {\em Mathematics of Computation}, 67(224):1577--1590, 1998.

\bibitem{reed1980methods}
Michael Reed and Barry Simon.
\newblock {\em Methods of modern mathematical physics: Functional analysis}, volume~1.
\newblock Gulf Professional Publishing, 1980.

\bibitem{qi2021survey}
Shuren Qi, Yushu Zhang, Chao Wang, Jiantao Zhou, and Xiaochun Cao.
\newblock A survey of orthogonal moments for image representation: Theory, implementation, and evaluation.
\newblock {\em ACM Computing Surveys (CSUR)}, 55(1):1--35, 2021.

\bibitem{katznelson2004introduction}
Yitzhak Katznelson.
\newblock {\em An introduction to harmonic analysis}.
\newblock Cambridge University Press, 2004.

\bibitem{mallat2002theory}
Stephane~G Mallat.
\newblock A theory for multiresolution signal decomposition: the wavelet representation.
\newblock {\em IEEE transactions on pattern analysis and machine intelligence}, 11(7):674--693, 2002.

\bibitem{kelley1993fast}
Brian~T Kelley and Vijay~K Madisetti.
\newblock The fast discrete radon transform. i. theory.
\newblock {\em IEEE Transactions on Image Processing}, 2(3):382--400, 1993.

\bibitem{flusser2006rotation}
Jan Flusser and Tom Suk.
\newblock Rotation moment invariants for recognition of symmetric objects.
\newblock {\em IEEE Transactions on Image Processing}, 15(12):3784--3790, 2006.

\bibitem{cohen2015approximation}
Albert Cohen and Ronald DeVore.
\newblock Approximation of high-dimensional parametric pdes.
\newblock {\em Acta Numerica}, 24:1--159, 2015.

\bibitem{lecun1998mnist}
Yann LeCun, L{\'e}on Bottou, Yoshua Bengio, and Patrick Haffner.
\newblock Gradient-based learning applied to document recognition.
\newblock {\em Proceedings of the IEEE}, 86(11):2278--2324, 1998.

\bibitem{kermany2018chestxray}
Daniel~S Kermany, Michael Goldbaum, Wenjia Cai, et~al.
\newblock Identifying medical diagnoses and treatable diseases by image-based deep learning.
\newblock {\em Cell}, 172(5):1122--1131.e9, 2018.

\bibitem{krizhevsky2009learning}
Alex Krizhevsky and Geoffrey Hinton.
\newblock Learning multiple layers of features from tiny images.
\newblock Technical report, Technical Report, University of Toronto, 2009.

\bibitem{zernike1934diffraction}
Frits Zernike.
\newblock Diffraction theory of the knife-edge test and its improved form, the phase-contrast method.
\newblock {\em Monthly Notices of the Royal Astronomical Society}, 94:377--384, 1934.

\bibitem{teh1988image}
Chin-Huat Teh and Roland~T. Chin.
\newblock On image analysis by the method of moments.
\newblock {\em IEEE Transactions on Pattern Analysis and Machine Intelligence}, 10(4):496--513, 1988.

\end{thebibliography}

\vspace*{-1cm}
\begin{IEEEbiography}[{\includegraphics[width=1in,height=1.25in,clip,keepaspectratio]{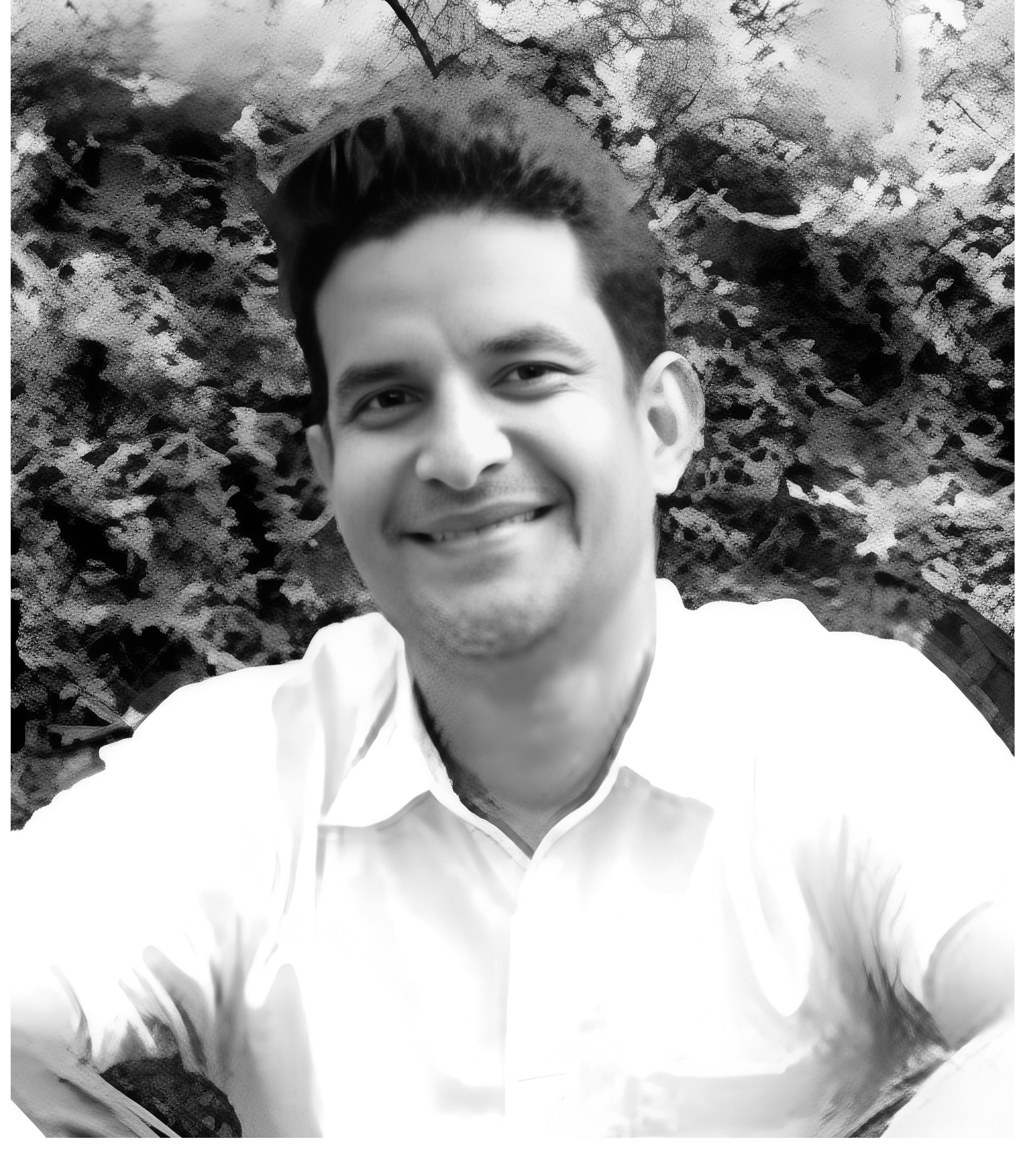}}]{Satya P. Singh}
(Member, IEEE) received his Ph.D. in Electrical Engineering with specialization in Artificial Intelligence and Medical Imaging, M.E. in Electronics Engineering from YMUST, Faridabad, and B.E. in Electronics and Telecommunication Engineering from IETE, New Delhi. He is an Associate Professor in the Department of Electronics and Communication Engineering, Netaji Subhas University of Technology (NSUT), New Delhi.  
Previously, he was a Postdoctoral Research Fellow at Nanyang Technological University, Singapore (2018–2021) and an Assistant Professor in India (2009–2018). His research interests include biomedical imaging, artificial intelligence, deep learning, and pattern recognition.  
Dr. Singh serves as an Associate Editor for \textit{Measurement}, \textit{Measurement: Sensors}, and \textit{Measurement: Food} (Elsevier).
\end{IEEEbiography}

\vspace*{-1cm}

\begin{IEEEbiography}[{\includegraphics[width=1in,height=1.25in,clip,keepaspectratio]{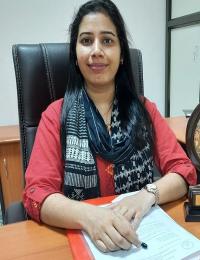}}]{Rashmi Chaudhry}
Dr. Chaudhry is an assistant professor at Computer Science and Engineering Department, Netaji Subhas University of Technology, New Delhi. She has done MTech and PhD from ISM(IIT) Dhanbad and ABV-IIITM, Gwalior respectively. Prior to joining NSUT, Delhi, she was Assistant Professor in Dr. SPM IIIT Naya Raipur, India. Her area of research includes Computer Networks, Wireless Sensor Networks, Internet of Things and Intelligent Transportation Networks. She has published research papers in various reputed journals and conference proceedings.
\end{IEEEbiography}

\vspace*{-1cm}

\begin{IEEEbiography}[{\includegraphics[width=1in,height=1.25in,clip,keepaspectratio]{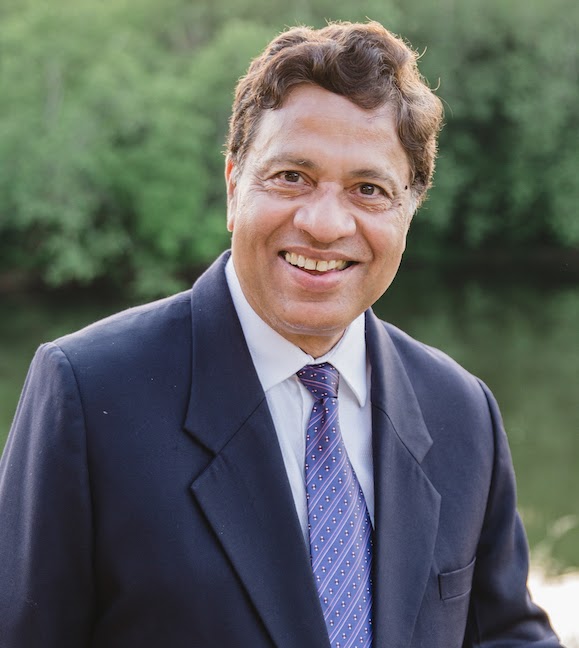}}]{Anand Srivastava}
Prof. Anand Srivastava is Vice Chancellor at Netaji Subhas University of Technology since November 2023. Previously, he was Professor in the Department of Electronics and Communication Engineering and Director of the Incubation Center at IIIT Delhi, and Adjunct Professor at the Bharti School of Telecom Technology, IIT Delhi. He has also served as Dean and Professor at IIT Mandi.  
Earlier, he worked as a solution architect at Alcatel-Lucent-Bell Labs and spent nearly 20 years at the Center for Development of Telematics (CDOT), where he led national-level telecom projects in security, network management, intelligent networks, access networks, and optical technologies. He conducted research at the Photonics Research Lab, Nice, France, and contributed to ITU-T optical networking standards. Currently, he drives VLC/LiFi standardization under TSDSI.  
His research interests include optical networks, vehicle-to-vehicle communications, FiWi architectures, and visible light communications. He received the "Distinguished Research Award" from IIIT Delhi and the "Outstanding Technical Contribution" award from TSDSI in 2023.
\end{IEEEbiography}
\vspace*{-1cm}
\begin{IEEEbiography}[{\includegraphics[width=1in,height=1.25in,clip,keepaspectratio]{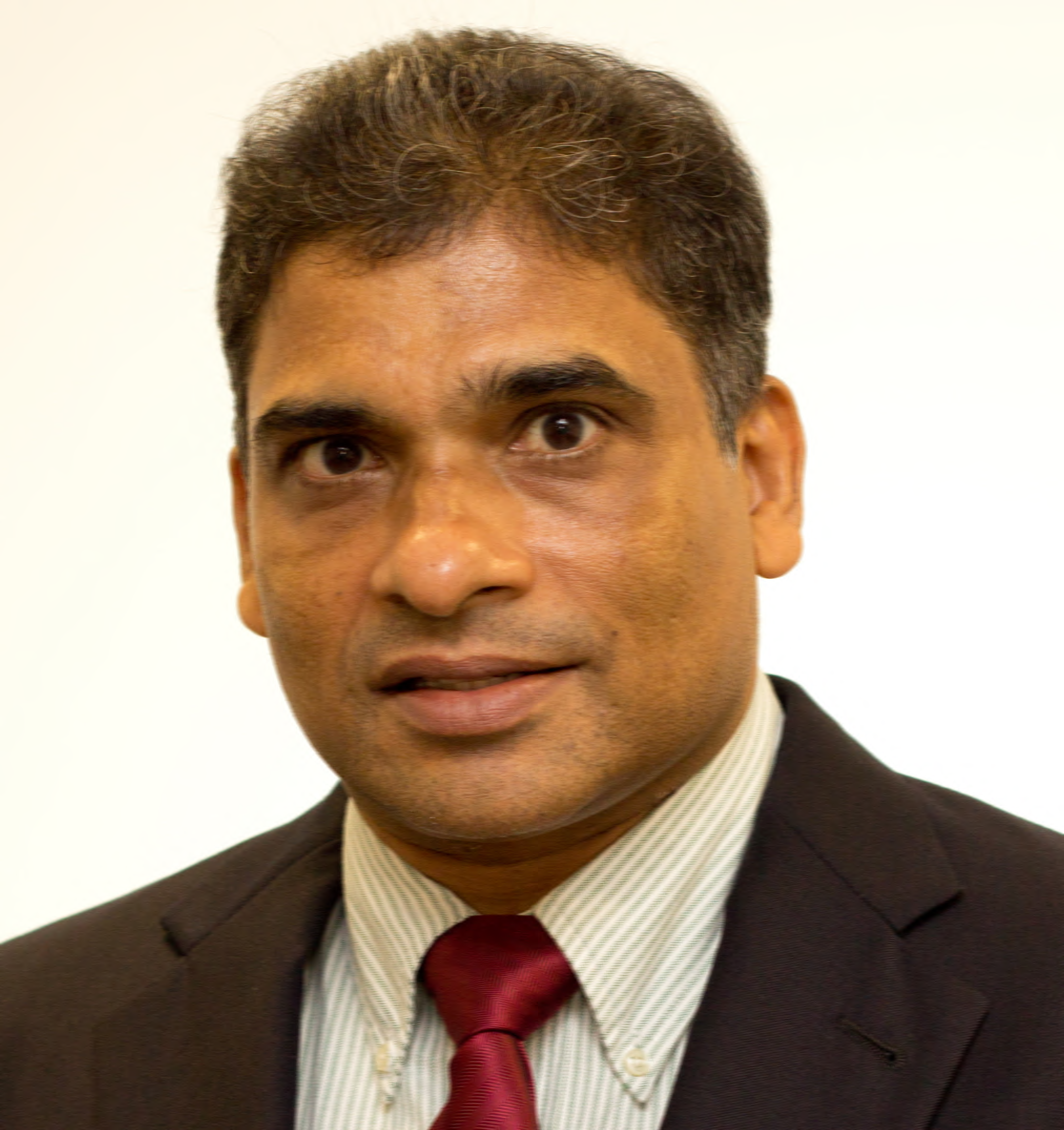}}]{Jagath Rajapakse}
(Fellow, IEEE) is Professor of Data Science at the College of Computing and Data Science, Nanyang Technological University (NTU), Singapore. He received the B.Sc. degree in Electronics and Telecommunication Engineering from the University of Moratuwa, Sri Lanka, and the M.S. and Ph.D. degrees in Electrical and Computer Engineering from the University at Buffalo, USA.  
He has held visiting positions at the Max-Planck Institute of Cognitive and Brain Sciences, Germany, the National Institute of Mental Health, USA, and the Department of Biological Engineering, Massachusetts Institute of Technology (MIT), USA. His research interests include explainable and generative AI, brain imaging, and computational and systems biology, with applications in brain disease diagnosis and drug discovery. He has authored over 300 peer-reviewed research articles.  
Prof. Rajapakse serves as Editor of \textit{Engineering Applications of Artificial Intelligence} and has previously served as Associate Editor for several IEEE Transactions. He was a Fulbright Scholar and was elevated to IEEE Fellow in 2012 for his contributions to brain image analysis.
\end{IEEEbiography}
\vspace*{10cm}

\end{document}